\documentclass{article} 
\usepackage{iclr2027_conference,times}

\usepackage{amsmath,amsfonts,bm}

\def\eqref#1{equation~\ref{#1}}

\def\1{\bm{1}}

\DeclareMathAlphabet{\mathsfit}{\encodingdefault}{\sfdefault}{m}{sl}
\SetMathAlphabet{\mathsfit}{bold}{\encodingdefault}{\sfdefault}{bx}{n}

\usepackage{hyperref}
\usepackage{url}
\usepackage{booktabs}
\usepackage{wrapfig}
\usepackage{multirow}
\usepackage{graphicx}
\usepackage{placeins}
\usepackage{amsmath,amssymb,amsthm}
\newtheorem{proposition}{Proposition}

\title{MercerFlow: Flow Matching in a Kernel-Induced Latent Space for Probabilistic Forecasting}

\author{Ilya Kuleshov\thanks{Equal contribution.}\, \thanks{\texttt{i.kuleshov@applied-ai.ru}} , Egor Serov\footnotemark[1] , Alexey Zaytsev \\
Applied AI Institute \\
Moscow, Russia
}

\iclrfinalcopy 

\begin{document}

\maketitle

\thispagestyle{fancy}
\lhead{Preprint. Under review.} 

\begin{abstract}
Recent work has shown that probabilistic flow matching for time series forecasting benefits from a data-matched prior.
The resulting prior introduces local correlations, which a sequential architecture usually absorbs: a recurrent neural network (RNN), a structured state-space model (S4), or a Transformer.
However, such a backbone costs GPU memory and time per epoch.
A cheaper alternative is MLP-based latent-space flow matching: embed the time series via an invertible map to a single latent vector and learn the flow there, so a tabular MLP can treat the series as a set of features.
The relationship between the prior and the choice of linear latent map is understudied in conditional flow matching (CFM) forecasting, yet we found it strongly affects performance.
Fixed transforms such as Fourier or discrete cosine (DCT) are only well-conditioned for Ornstein--Uhlenbeck priors, while a principal-component (PCA) map fit to the data is a strong but training-set-dependent reference sensitive to train--test shift.
Instead, we propose to use the Mercer eigenbasis of the prior kernel: it diagonalises the centred covariance exactly, decouples from training data, and adapts to non-stationary and periodic priors.
On five GluonTS benchmarks (ETTh1, ETTh2, Weather, Electricity, Traffic) under a shared protocol with TSFlow, the resulting MLP matches or beats it on CRPS at about $4.7\times$ less training memory and $3.5\times$--$4.4\times$ less time per epoch.
\end{abstract}

\section{Introduction}
The usual approach to probabilistic forecasting is to devise an algorithm which samples from a predictive distribution over future windows given the observed past.
A sampler that is expressive, while staying cheap to train and cheap to run, is the practical target.
Flow matching~\citep{lipman2023flow,tong2024improving} is a prominent modern way to build such samplers: one learns a velocity field along a probability path from a tractable prior to the data, and training is ordinary regression onto a known interpolant.
Since each forecast requires numerically integrating this velocity during inference, the cost at test time is dominated by the backbone that evaluates the field; cheapness at training time matters too, since the backbone also dictates GPU memory and time per epoch.
If the prior already matches the data, the flow starts closer to the target and needs less capacity to finish the trip.
TSFlow~\citep{kollovieh2025tsflow} commits to a Gaussian-process (GP) prior on the forecast window for that reason.
However, under that prior, neighbouring time coordinates stay correlated, so a global multilayer perceptron (MLP) spends capacity on local smoothness the GP already encodes.
TSFlow absorbs those correlations with a bidirectional state-space backbone~\citep{gu2022efficiently}, which is accurate on long smooth series but expensive in memory and time per epoch.

Orthonormal projections are a classical way to decorrelate coordinates in a latent space~\citep{mercer1909functions,karhunen1947uber,gabor1946theory}.
Several recent generative models run denoising or score matching in orthonormal latents: DCTdiff~\citep{ning2025dctdiff} trains image diffusion in discrete cosine coordinates, Crabb{\'e} et al.~\citep{crabbe2024fourier} dualise time-series diffusion to the Fourier domain, and Phillips et al.~\citep{phillips2022spectral} score functional data along truncated Mercer eigenfunctions.
However, when we commit to a GP prior, as in TSFlow, the latent map must also decorrelate the resulting structure—not just the correlations in the data itself.
To address this, we conduct a controlled comparison of five linear codecs for projection on a shared architecture and matched prior: Fourier (discrete Fourier transform), discrete cosine (DCT), principal components (PCA), the identity (Raw), and Mercer-based eigenbasis.

Mercer's theorem supplies the orthonormal eigenbasis of the kernel one has already committed to~\citep{mercer1909functions,karhunen1947uber}, so the coefficients are uncorrelated under the corresponding GP.
Thus, that decorrelation is exact for the centred prior and approximate for the conditional law and for the interpolant on which the network trains.
Across codecs, Mercer is the most versatile data-free choice: it works on data which requires a non-OU prior (Fourier and DCT best diagonalise OU-like priors, which are not always optimal for the data) and where PCA suffers from the train--test shift since it fits only the training path (§\ref{sec:why-mercer}).

That said, all five codecs are orthonormal projections, so a rotation-equivariant optimiser with an MSE loss should treat them as equivalent reparameterisations~\citep{nesterov2004introductory,zhang2025adamrotation}.
However, that equivalence is optimiser-specific: it holds for Euclidean SGD (Proposition~\ref{prop:sgd-invariance}), but fails under Adam because of the first-layer input geometry.
Write $A=\mathbb{E}[\mathbf{x}\mathbf{x}^\top]$ for a first-layer input $\mathbf{x}$ second-moment matrix and $\kappa(A)=\lambda_{\max}(A)/\lambda_{\min}(A)$ for the condition number of its spectrum.
Orthonormal codecs leave the spectrum and $\kappa(A)$ unchanged, so Euclidean SGD cannot prefer one codec over another.
Under Adam, however, per-coordinate rescaling is modelled by the Jacobi preconditioner $D=\mathrm{diag}(A)^{-1/2}$, after which the normalised second-moment matrix is $DAD$ with the geometry conditioned by $\kappa(DAD)$.
The rotation leaves $\kappa(A)$ fixed but changes $\mathrm{diag}(A)$, which is what Jacobi reads, so $\kappa(DAD)$ is codec-dependent.
Empirically, the decorrelating codecs exhibit both lower
$\kappa(DAD)$ and better forecasting accuracy than Raw under
Adam on challenging datasets, while their advantage
disappears under SGD.
Our experiments therefore examine how codec choice interacts with the optimiser and whether MercerFlow offers a favourable balance of predictive accuracy and computational cost relative to TSFlow and other baselines.

\begin{figure*}[t]
\centering
\includegraphics[width=\textwidth]{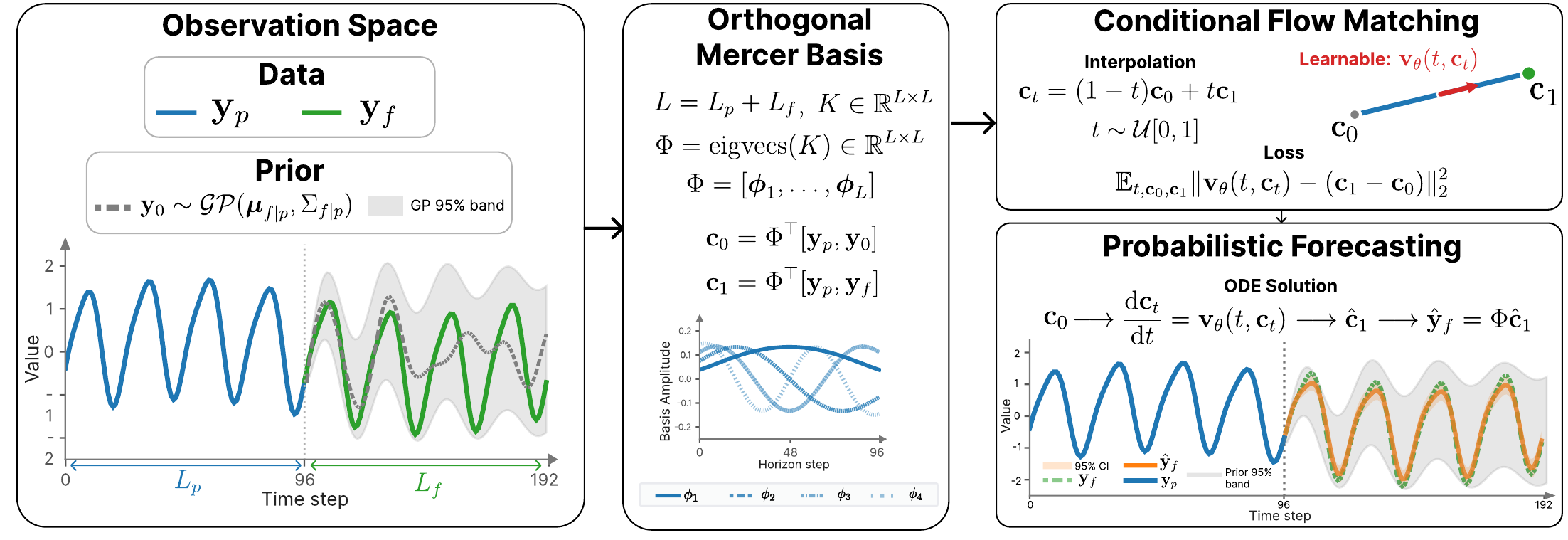}
\caption{\textbf{Overview of \textsc{MercerFlow}.} The joint context and horizon window is mapped into the decorrelated Mercer eigenbasis of the GP prior. A lightweight MLP learns conditional flow matching directly in this latent space, enabling fast and scalable probabilistic forecasting without sequential backbones.}
\label{fig:pipeline}
\end{figure*}

\paragraph{Contributions.}
Our contributions are as follows:
\begin{itemize}
\item A controlled five-codec study of conditional flow matching under matched architecture, prior, and protocol: Mercer, Fourier, DCT, PCA, and Raw on five GluonTS benchmarks with paired $t$-test ranks.
\item A diagnostic of codec gaps under Adam versus SGD. First-layer $\kappa(\mathrm{DAD})$ separates Raw from the decorrelating codecs on hard sets under Adam, while the gaps vanish when the models are trained with SGD.
\item \textsc{MercerFlow}: conditional flow matching carried out directly in the Mercer / Karhunen--Lo\`eve eigenbasis of the forecasting GP prior.
Unlike fixed transforms such as DCT and Fourier, the basis adapts directly to non-stationary and periodic priors, providing a data-free representation that remains robust across kernels without fitting the training path.
\item An empirical comparison against TSFlow on multiple benchmarks.
The MLP codecs are about $4.7\times$ cheaper in peak GPU memory and $3.5\times$--$4.4\times$ cheaper in time per epoch at matched batch size, making MLP flow matching in linearly projected latents a practical alternative to sequential S4 backbones.
\end{itemize}

\section{Method}

We study probabilistic time series forecasting, aiming to model the conditional distribution of a future window given the observed past. 
Following the channel-independent setting of
TSFlow~\citep{kollovieh2025tsflow}, each channel is treated as a separate univariate series.
Let $L_{\mathrm{p}}$ and $L_{\mathrm{f}}$ denote the context (past) and forecast horizon (future) lengths, respectively, yielding a total sequence length $L = L_{\mathrm{p}} + L_{\mathrm{f}}$.
We define the regular discrete time grid $\mathcal{S} = \{0, \dots, L-1\}$, partitioned into the observed past $\mathcal{S}_{\mathrm{p}} = \{0, \dots, L_{\mathrm{p}}-1\}$ and the future forecast horizon $\mathcal{S}_{\mathrm{f}} = \{L_{\mathrm{p}}, \dots, L-1\}$.
The observed past series is denoted by $\mathbf{y}_{\mathrm{p}} \in \mathbb{R}^{L_{\mathrm{p}}}$, the true future window by $\mathbf{y}_{\mathrm{f}} \in \mathbb{R}^{L_{\mathrm{f}}}$, and their joint concatenation by $\mathbf{y} = [\mathbf{y}_{\mathrm{p}}, \mathbf{y}_{\mathrm{f}}] \in \mathbb{R}^L$.

\paragraph{MercerFlow overview.}
Figure~\ref{fig:pipeline} summarises three stages of \textsc{MercerFlow}.
First, we condition a Gaussian-process prior on the observed past and
sample an initial future window (Subsection~\ref{sec:gp-prior}).
Second, we concatenate this sample with the past and encode the joint window in the orthonormal eigenbasis of the same kernel matrix (Subsection~\ref{sec:codec-set}).
Third, during training, a lightweight MLP learns the velocity along
linear interpolants between the encoded source and target windows
(Subsection~\ref{sec:vector-field}).
Finally, at inference time, we integrate the learned velocity from a
newly sampled source, apply the inverse transform, and extract the
forecast horizon (Subsection~\ref{sec:vector-field}).
Repeating this procedure produces samples from the predictive distribution.

\subsection{Conditional GP Prior}
\label{sec:gp-prior}

Let $k: \mathcal{S} \times \mathcal{S} \to \mathbb{R}$ be a positive-definite covariance kernel, e.g., Ornstein--Uhlenbeck (OU), radial basis function (RBF), or periodic, with length scale $\ell > 0$ and diagonal jitter $\varepsilon > 0$. 
In our primary benchmark runs, we adopt the stationary exponential (OU) kernel:
\begin{equation}
k(s, s') = \exp\left(-\frac{|s - s'|}{\ell}\right), \quad s, s' \in \mathcal{S}.
\label{eq:kernel_def}
\end{equation}
Evaluating $k$ across all pairs of temporal coordinates in $\mathcal{S}$ defines the joint covariance matrix $K \in \mathbb{R}^{L \times L}$, with entries $K_{ij} = k(i, j) + \varepsilon I[i = j]$. 
Partitioning $K$ according to the past and future index sets $\mathcal{S}_{\mathrm{p}}$ and $\mathcal{S}_{\mathrm{f}}$ yields:
\begin{equation}
K = \begin{bmatrix}
K_{\mathrm{pp}} & K_{\mathrm{pf}} \\
K_{\mathrm{fp}} & K_{\mathrm{ff}}
\end{bmatrix},
\label{eq:gram_blocks}
\end{equation}
where $K_{\mathrm{pp}} \in \mathbb{R}^{L_{\mathrm{p}} \times L_{\mathrm{p}}}$, $K_{\mathrm{ff}} \in \mathbb{R}^{L_{\mathrm{f}} \times L_{\mathrm{f}}}$, and $K_{\mathrm{fp}} = K_{\mathrm{pf}}^\top \in \mathbb{R}^{L_{\mathrm{f}} \times L_{\mathrm{p}}}$. 
Conditioned on the observed context $\mathbf{y}_{\mathrm{p}}$, the observation-space prior over the future window $\mathbf{y}_0 \in \mathbb{R}^{L_{\mathrm{f}}}$ is given by the standard Gaussian Process posterior:
\begin{equation}
\mathbf{y}_0 \sim \mathcal{N}(\boldsymbol{\mu}, \boldsymbol{\Sigma}), \qquad
\boldsymbol{\mu} = K_{\mathrm{fp}} K_{\mathrm{pp}}^{-1} \mathbf{y}_{\mathrm{p}}, \qquad
\boldsymbol{\Sigma} = K_{\mathrm{ff}} - K_{\mathrm{fp}} K_{\mathrm{pp}}^{-1} K_{\mathrm{pf}}.
\label{eq:gp-prior}
\end{equation}
We select the prior length scale $\ell$ via validation-likelihood maximisation per dataset (Table~\ref{tab:ou_lengthscales}, \S\ref{sec:experiments}).
The diagonal jitter $\varepsilon$ provides numerical stability but cannot substitute for basis rotation: as shown in Appendix~\ref{app:jitter-ablation}, Raw's condition number remains near $10^5$ across all $\varepsilon \le 10^{-2}$, while extreme regularisation ($\varepsilon = 1$) degenerates the prior into white noise and degrades both Mercer and Raw equally.

\subsection{Linear Codec on the Joint Window}
\label{sec:codec-set}

To make the observed past available to the velocity field, we apply a single linear codec to the joint context and future window.
The source window concatenates $\mathbf{y}_{\mathrm{p}}$ with a sample $\mathbf{y}_0$ from the conditional GP prior, while the
target window combines the same context with the observed
future $\mathbf{y}_{\mathrm{f}}$.
Both windows are projected into a shared coefficient space using
an orthonormal matrix $\Phi \in \mathbb{R}^{L \times L}$,
with encoding and decoding defined by:
\begin{equation}
\mathbf{c} = \Phi^\top \mathbf{y}, \qquad \mathbf{y} = \Phi \mathbf{c}.
\label{eq:codec}
\end{equation}
Since the columns of $\Phi$ are orthonormal, $\Phi \Phi^\top = I_L$, and this map is an exact isometry.
The source prior state is $\mathbf{c}_0 = \Phi^\top [\mathbf{y}_{\mathrm{p}}, \mathbf{y}_0]$ and the target state is $\mathbf{c}_1 = \Phi^\top [\mathbf{y}_{\mathrm{p}}, \mathbf{y}_{\mathrm{f}}]$.
The five codecs described below differ only in how they construct the basis vectors $\boldsymbol{\phi}_i$ of the matrix $\Phi = [\boldsymbol{\phi}_1, \dots, \boldsymbol{\phi}_L]$.

\textbf{Identity (Raw).}
This codec keeps the series in observation coordinates with $\Phi = I_L$, which yields $\mathbf{c} = \mathbf{y}$.
Each coordinate represents an ambient time step.
It acts as an unrotated baseline and matches the time-domain setup of TSFlow, but uses an MLP rather than an S4 backbone.

\textbf{Mercer (Ours).}
The basis columns $\boldsymbol{\phi}_i$ are the orthonormal eigenvectors of the kernel matrix $K$:
\begin{equation}
K = \Phi \Lambda \Phi^\top, \qquad \Lambda = \operatorname{diag}(\lambda_1, \dots, \lambda_L).
\label{eq:mercer}
\end{equation}
Eigenvalues are sorted in descending order.
The first row of $\Phi$ is constrained to be non-negative to fix signs.
By Mercer's theorem and the Karhunen--Lo\`eve expansion, if $\mathbf{y} \sim \mathcal{N}(\mathbf{0}, K)$, then $\operatorname{Cov}(\mathbf{c}) = \Lambda$~\citep{rasmussen2006gaussian}.
Appendix~\ref{app:jacobi} gives the exact derivation. Therefore, latent coordinates are uncorrelated under the prior.
This decorrelation is exact for the prior and provides an effective preconditioning proxy for the conditional law.

\textbf{Fourier.}
This codec implements a real discrete Fourier transform on the time grid $s \in \mathcal{S}$~\citep{gabor1946theory,baraniuk1995unitary}.
The first column is the constant direct current (DC) mode $L^{-1/2}$.
Higher harmonic frequencies $m = 1, 2, \dots$ provide cosine and sine pairs:
\begin{equation}
\begin{aligned}
\phi_m^{\cos}(s)
= \sqrt{\frac{2}{L}}\,\cos\Bigl(\frac{2\pi m s}{L}\Bigr), 
\phi_m^{\sin}(s)
= \sqrt{\frac{2}{L}}\,\sin\Bigl(\frac{2\pi m s}{L}\Bigr).
\end{aligned}
\label{eq:fourier-modes}
\end{equation}
When $L$ is even, the final column is the Nyquist mode $L^{-1/2}\cos(\pi s)$.
The transform $\mathbf{c} = \Phi^\top \mathbf{y}$ expresses the
series in this real orthonormal Fourier basis, with each coefficient
$c_i = \boldsymbol{\phi}_i^\top \mathbf{y}$ corresponding to
one Fourier mode.

\textbf{DCT.}
This codec uses an orthonormal type-II discrete cosine transform, ordered from lowest to highest frequency.
Evaluated at time steps $s \in \mathcal{S}$, the basis columns are:
\begin{equation}
\phi_0^{\mathrm{DCT}}(s)
= L^{-1/2}, \qquad
\phi_m^{\mathrm{DCT}}(s)
= \sqrt{\frac{2}{L}}\,
   \cos\left(\frac{\pi m(s+1/2)}{L}\right),
   \qquad m=1,\ldots,L-1.
\label{eq:dct-modes}
\end{equation}
With $L^{-1/2}$ and $\sqrt{2/L}$ normalisation, the modes form an orthonormal basis.
Unlike Fourier, the DCT does not enforce periodic boundaries.
It provides an asymptotic Karhunen--Lo\`eve basis for stationary Markovian and OU processes~\citep{ahmed1974discrete,gray2006toeplitz}.

\textbf{PCA.}
The basis columns $\boldsymbol{\phi}_i$ are the eigenvectors of the empirical sample covariance matrix computed across all training windows $\{\mathbf{y}^{(k)}\}_{k=1}^S$~\citep{jolliffe2002principal}:
\begin{equation}
\widehat{\Sigma}_{\mathrm{train}} = \frac{1}{S} \sum_{k=1}^S (\mathbf{y}^{(k)} - \bar{\mathbf{y}})(\mathbf{y}^{(k)} - \bar{\mathbf{y}})^\top, \qquad \widehat{\Sigma}_{\mathrm{train}} = \Phi \Lambda \Phi^\top,
\label{eq:pca}
\end{equation}
here $\bar{\mathbf{y}}$ is the empirical mean of the training sequences.
Eigenvalues are sorted in descending order with the same sign convention as Mercer.
Centering is used solely to construct $\widehat{\Sigma}_{\mathrm{train}}$, while the latent projection $\mathbf{c} = \Phi^\top \mathbf{y}$ remains strictly linear.
Because $\Phi$ is fitted to the training sample, PCA is data-dependent and sensitive to train--test distribution shifts.

\subsection{Flow Matching in Coefficient Space}
\label{sec:vector-field}
Flow matching~\citep{lipman2023flow} learns a time-dependent velocity field that transports a source distribution to the data.
Standard formulations often require minibatch optimal transport to straighten trajectories originating from unstructured Gaussian noise~\citep{tong2024improving}.
In contrast, we adopt TSFlow~\citep{kollovieh2025tsflow} and use the conditional GP prior~\eqref{eq:gp-prior} as the source distribution.
Because this prior already captures temporal history and local smoothness, independent pairing between source and target yields straight paths without optimal transport overhead.

We instantiate this probability path directly in the joint coefficient space:
\begin{equation}
\mathbf{c}_t = (1 - t)\mathbf{c}_0 + t\mathbf{c}_1, \qquad
\mathbf{u} = \mathbf{c}_1 - \mathbf{c}_0,
\label{eq:path}
\end{equation}
with flow time $t \sim \mathcal{U}[0, 1]$, source $\mathbf{c}_0$, and target $\mathbf{c}_1$.
Since the historical context is shared, the target displacement simplifies to $\mathbf{u} = \Phi^\top [\mathbf{0}, \mathbf{y}_{\mathrm{f}} - \mathbf{y}_0]$.
We train an MLP velocity field $\mathbf{v}_\theta(t, \mathbf{c}_t)$ by minimizing the mean squared regression error:
\begin{equation}
\mathcal{L}(\theta) = \mathbb{E}_{t, \mathbf{c}_0, \mathbf{c}_1} \left\| \mathbf{v}_\theta(t, \mathbf{c}_t) - \mathbf{u} \right\|_2^2.
\label{eq:loss}
\end{equation}
Because $\Phi$ is an orthogonal isometry ($\Phi \Phi^\top = I_L$), it preserves the Euclidean norm: $\|\mathbf{u}\|_2 = \|\mathbf{y}_{\mathrm{f}} - \mathbf{y}_0\|_2$.
Consequently, the training loss magnitude is identical across all five codecs (Appendix~\ref{app:sgd}).

To preserve input-layer geometry, the MLP injects AdaLN time modulation only after the first linear map (Appendix~\ref{app:adaln}).
During inference, we sample $\mathbf{y}_0 \sim \mathcal{N}(\boldsymbol{\mu}, \boldsymbol{\Sigma})$, project the joint state $\mathbf{c}_0$, and integrate $\mathbf{v}_\theta$ from $t = 0$ to $1$ using an Euler ODE solver to obtain the terminal state $\hat{\mathbf{c}}_1$.
The predicted horizon $\hat{\mathbf{y}}_{\mathrm{f}}$ is then extracted from the reconstructed sequence $\hat{\mathbf{y}} = \Phi \hat{\mathbf{c}}_1$.

\subsection{Why the Rotation Can Matter: Adam's Diagonal}
\label{sec:theory}

At full order, corresponding to a square orthonormal basis $\Phi \in \mathbb{R}^{L \times L}$ on the ambient sequence window, Mercer, Fourier, and the identity baseline parameterise the identical Euclidean interpolant in three distinct orthonormal coordinate systems.

\begin{proposition}[SGD is invariant to the choice of orthonormal codec]
\label{prop:sgd-invariance}
Let $\Phi \in \mathbb{R}^{L \times L}$ be a square orthogonal matrix defined on the ambient observation space.
Consider the velocity field $\mathbf{v}_\theta$ parameterized by $\theta = \{W_1, \mathbf{b}_1, W_2, \mathbf{b}_2, \theta_\psi\}$:
\begin{equation}
\mathbf{v}_\theta(t, \mathbf{z}) = W_2 \, \psi\bigl(W_1 \mathbf{z} + \mathbf{b}_1, \, t; \, \theta_\psi\bigr) + \mathbf{b}_2,
\label{eq:prop1_field}
\end{equation}
where $\psi(\cdot, t; \, \theta_\psi)$ aggregates all internal hidden layers along with AdaLN time conditioning, acting entrywise on its first argument with parameters $\theta_\psi$ independent of $\Phi$.

Let $\theta_I = \{W_1, \mathbf{b}_1, W_2, \mathbf{b}_2, \theta_\psi\}$ denote the parameters of the identity network, and define the conjugated parameter set $\theta_\Phi = \{W_1^\Phi, \mathbf{b}_1^\Phi, W_2^\Phi, \mathbf{b}_2^\Phi, \theta_\psi^\Phi\}$ of the codec-$\Phi$ network by
\begin{equation}
W_1^\Phi = W_1 \Phi, \qquad 
W_2^\Phi = \Phi^\top W_2, \qquad 
\mathbf{b}_2^\Phi = \Phi^\top \mathbf{b}_2, \qquad 
\mathbf{b}_1^\Phi = \mathbf{b}_1, \qquad
\theta_\psi^\Phi = \theta_\psi.
\label{eq:param_conjugation}
\end{equation}
Then the velocity field satisfies $\mathbf{v}_{\theta_\Phi}(t, \Phi^\top \mathbf{z}) = \Phi^\top \mathbf{v}_{\theta_I}(t, \mathbf{z})$, and the squared conditional flow matching loss~\eqref{eq:loss} is preserved identically:
\begin{equation}
\|\mathbf{v}_{\theta_\Phi} - \mathbf{u}\|_2^2 = \|\mathbf{v}_{\theta_I} - (\mathbf{z}_1 - \mathbf{z}_0)\|_2^2,
\label{eq:loss_equivariance}
\end{equation}
where $\mathbf{u} = \Phi^\top(\mathbf{z}_1 - \mathbf{z}_0)$.
Under rotation-invariant parameter initialisations, Euclidean gradient descent and Euclidean momentum schemes, including the heavy-ball method and Nesterov accelerated gradient, generate identical loss trajectories and identical decoded flows at every training iteration~\citep{nesterov2004introductory,zhang2025adamrotation}.
\end{proposition}
The proof is provided in Appendix~\ref{app:sgd}.
An empirical validation using Nesterov SGD at a matched computational budget is reported for ETTh1 and ETTh2 in Table~\ref{tab:joint_sgd_full} of Appendix~\ref{app:sgd-control}.

The equivalence established in Proposition~\ref{prop:sgd-invariance} is strictly Euclidean, as standard stochastic gradient descent applies isotropic updates across all coordinate axes.
Modern probabilistic forecasting models are instead routinely optimized with adaptive algorithms such as AdamW~\citep{kingma2015adam}.
These optimizers maintain an exponential moving average of squared coordinate-wise gradients to rescale parameter updates individually, thereby dampening oscillations along steep directions and accelerating progress along plateaus.
Because this coordinate-wise scaling is defined relative to the canonical parameter axes, it is not rotation-equivariant.
Rotating the input-space basis fundamentally alters Adam's trajectory, whereas SGD's trajectory undergoes a simple linear conjugation~\citep{zhang2025adamrotation,xie2025adamlinf}.
Consequently, when training the velocity field $\mathbf{v}_\theta$ of Section~\ref{sec:vector-field} with AdamW, optimisation dynamics are governed not by the raw input Gram matrix, but by its geometry following diagonal preconditioning.

Let $\mathbf{x} \in \mathbb{R}^L$ denote the input vector to the first linear layer, and let $A = \mathbb{E}[\mathbf{x}\mathbf{x}^\top]$ be its uncentred second-moment matrix.
For the linear mapping $\mathbf{h} = W \mathbf{x}$, the matrix $A$ determines the curvature of the corresponding linear subproblem.
The extremal directions of stretch are given by the eigenvectors of $A$, and their relative ratio is quantified by the condition number
\begin{equation}
\kappa(A) = \frac{\lambda_{\max}(A)}{\lambda_{\min}(A)}.
\label{eq:kappa}
\end{equation}
As \citet{nesterov2004introductory} show, gradient descent on a quadratic objective has a convergence rate governed directly by $\kappa(A)$.
Because an orthogonal transformation leaves the spectrum of $A$ invariant, substituting $A$ with $Q^\top A Q$ preserves $\kappa(A)$ and cannot accelerate Euclidean gradient descent.

In contrast, the first-layer dynamics of Adam deviate substantially from Euclidean optimisation.
Under the Kronecker-factored curvature model of \citet{martens2015kfacc}, the coordinate-wise second-moment accumulator decomposes into an output factor and an input factor given by $\operatorname{diag}(A)$.
The resulting coordinate rescaling corresponds to the classical Jacobi preconditioner $D = \operatorname{diag}(A)^{-1/2}$, producing the preconditioned input $\tilde{\mathbf{x}} = D \mathbf{x}$ and the rescaled curvature matrix $D A D$.
The optimisation of the first-layer weights is therefore dictated by $\kappa(D A D)$.
According to \citet{vandersluis1969}, the Jacobi preconditioner minimises the condition number among all diagonal scalings up to a factor depending only on dimensionality.
When $A$ possesses small off-diagonal elements, $D A D$ approximates the identity matrix, effectively whitening the layer inputs.

\begin{proposition}[Jacobi preconditioning distinguishes linear codecs]
\label{prop:adam-jacobi}
Let the total window length be $n = L$, and let $T_n(f) \in \mathbb{R}^{n \times n}$ denote the symmetric Toeplitz covariance matrix generated by the spectral density function $f(\omega)$ of a stationary Ornstein--Uhlenbeck process with length scale $\ell$:
\begin{equation}
\bigl[T_n(f)\bigr]_{jk} = \frac{1}{2\pi} \int_{-\pi}^\pi f(\omega) \, e^{i(j-k)\omega} \, d\omega = \exp\left(-\frac{|j-k|}{\ell}\right).
\label{eq:toeplitz_def}
\end{equation}
Let $\mathbf{x} = \Phi^\top \mathbf{y}$ be the coefficient representation under the centred prior $\mathbf{y} \sim \mathcal{N}(\mathbf{0}, K)$ with $K = T_n(f) + \delta I_n$, and let $A = \mathbb{E}[\mathbf{x}\mathbf{x}^\top]$.
Under the Jacobi preconditioner $D = \operatorname{diag}(A)^{-1/2}$, the condition number $\kappa(D A D)$ satisfies:
\begin{enumerate}
    \item \textbf{Identity (Raw):} The matrix $A = T_n(f) + \delta I_n$ is Toeplitz with constant diagonal $1 + \delta$. The preconditioner satisfies $D = (1+\delta)^{-1/2} I_n$, yielding $\kappa(D A D) = \kappa(A)$, which grows monotonically with the prior length scale $\ell$.
    \item \textbf{Mercer:} Because the columns of $\Phi$ diagonalize the kernel matrix $K$, the second-moment matrix satisfies $A = \Lambda$, where $\Lambda$ is diagonal. Hence $D A D = I_n$, which yields the optimal condition number $\kappa(D A D) = 1$.
    \item \textbf{Fourier and DCT:} The respective transformation bases satisfy the asymptotic bounds
    \begin{equation}
    \kappa(D A D) \le 1 + \mathcal{O}\left(\frac{1}{\min_\omega f(\omega)}\right),
    \label{eq:fourier_dct_bound}
    \end{equation}
    where the bound loosens whenever $f(\omega)$ concentrates mass predominantly at low frequencies.
\end{enumerate}
\end{proposition}
The proof is given in Appendix~\ref{app:jacobi}, where Table~\ref{tab:jacobi_bounds} provides a structured overview of the second-moment structure and condition numbers across all five codecs.

On synthetic stationary OU paths where the data-generating kernel is known exactly, empirical $\kappa(DAD)$ matches this ordering and CRPS ties across codecs within $0.2\%$ (Appendix~\ref{app:synthetic-ou}), reproducing under analytical control the behavior observed on easy real datasets.

While an orthonormal change of basis cannot alter $\kappa(A)$, it modifies the diagonal entries $\operatorname{diag}(A)$ that govern Adam's scaling factors.
During training, the network input is not drawn from the centred prior $\mathbf{y}$, but represents the conditional interpolant $\mathbf{c}_t$.
Empirical evaluations of $\kappa(D A D)$ across datasets are detailed in Appendix~\ref{app:dataset-hardness} (Table~\ref{tab:joint_kdad_cond}) and summarized in Section~\ref{sec:why-mercer}.
Under AdamW optimisation, first-layer $\kappa(D A D)$ clearly separates the ill-conditioned Raw baseline from the decorrelating codecs on challenging benchmarks, whereas training with SGD eliminates performance discrepancies across codecs, in complete alignment with Proposition~\ref{prop:sgd-invariance}.

\section{Experiments}
\label{sec:experiments}

\textbf{Benchmark Setup and Protocol.} 
Experiments cover five standard time series benchmarks: ETTh1,
ETTh2, Weather, Electricity, and Traffic.
The context and forecast lengths are fixed at
$L_{\mathrm{p}} = L_{\mathrm{f}} = 192$, giving a joint window
of length $L = 384$, with channels treated independently.

Baselines include four alternative codecs (Raw, Fourier, DCT,
and PCA), the flow-based forecaster TSFlow, and the deterministic
forecasters DLinear, PatchTST, and DMamba.
All five codec variants use the same MLP architecture,
joint-window conditioning scheme, and GP prior.
Prior hyperparameters are selected by validation likelihood
(Appendix~\ref{app:dataset-kernels}).
Architecture and training details appear in
Appendix~\ref{app:experimental-setup}.

For each probabilistic model, evaluation uses $S = 100$ forecast
trajectories $\{\hat{\mathbf{y}}_{\mathrm{f}}^{(s)}\}_{s=1}^{S}$
per test window.
Normalised continuous ranked probability score (CRPS) measures
distributional forecast accuracy
\citep{alexandrov2020gluonts,kollovieh2025tsflow},
while mean absolute error (MAE) measures point-forecast accuracy.
For probabilistic models, the point forecast is the empirical
median of the generated samples.
Scores are aggregated over rolling test windows, with metric
definitions and aggregation details provided in
Appendix~\ref{sec:metrics}.

\subsection{Main Results: Forecasting Quality and Efficiency}
\label{sec:main-results}

Table~\ref{tab:joint_ranks} summarises overall prediction quality across all five benchmarks via paired two-sided $t$-test ranks, while Table~\ref{tab:efficiency} reports the computational footprint. Full per-dataset numerical CRPS and MAE distributions are reported in Appendix~\ref{app:full-crps-mae} within Tables~\ref{tab:main_scoreboard} and~\ref{tab:full_mae_appendix}.

\begin{table}[tbh]
\caption{\textbf{Predictive performance summary} across all five benchmarks (ETTh1, ETTh2, Weather, Electricity, Traffic). Rank $= 1 +$ number of methods strictly outperforming the model under a paired two-sided $t$-test on shared seeds ($p < 0.2$, lower score is better); a win denotes rank $1$.}
\label{tab:joint_ranks}
\vskip 0.05in
\begin{center}
\begin{small}
\setlength{\tabcolsep}{7pt}
\begin{tabular}{lcccccc}
\toprule
& \textbf{Ours} & \multicolumn{4}{c}{\textbf{Linear Codec Baselines}} & \textbf{Sequential} \\
\cmidrule(lr){2-2} \cmidrule(lr){3-6} \cmidrule(lr){7-7}
\textbf{Metric} & \textbf{Mercer} & \textbf{Fourier} & \textbf{DCT} & \textbf{PCA} & \textbf{Raw} & \textbf{TSFlow (S4)} \\
\midrule
Mean CRPS rank ($\downarrow$) & $\mathbf{1.00}$ & $1.20$ & $1.40$ & $1.80$ & $3.40$ & $5.80$ \\
\# CRPS wins ($\uparrow$)       & $\mathbf{5/5}$  & $4/5$  & $4/5$  & $3/5$  & $2/5$  & $0/5$  \\
\midrule
Mean MAE rank ($\downarrow$)  & $\mathbf{1.00}$ & $1.60$ & $1.40$ & $2.00$ & $3.20$ & $5.00$ \\
\# MAE wins ($\uparrow$)        & $\mathbf{5/5}$  & $3/5$  & $4/5$  & $3/5$  & $2/5$  & $1/5$  \\
\bottomrule
\end{tabular}
\end{small}
\end{center}
\vskip -0.1in
\end{table}

\begin{table}[tbh]
\caption{\textbf{Computational efficiency benchmark} at batch size 512. Comparison of per-epoch training time (seconds, $\downarrow$) and peak GPU memory (GB, $\downarrow$).}
\label{tab:efficiency}
\vskip 0.05in
\begin{center}
\begin{small}
\setlength{\tabcolsep}{8pt}
\begin{tabular}{lccccc}
\toprule
\textbf{Metric / Architecture} & \textbf{ETTh1} & \textbf{ETTh2} & \textbf{Weather} & \textbf{Electricity} & \textbf{Traffic} \\
\midrule
MLP Time (s, $\downarrow$)  & \textbf{0.67} & \textbf{0.69} & \textbf{1.80} & \textbf{0.93} & \textbf{0.68} \\
S4 Time (s, $\downarrow$)   & 2.50          & 2.40          & 7.70          & 4.10          & 2.60          \\
\midrule
MLP VRAM (GB, $\downarrow$) & \textbf{1.5}  & \textbf{1.5}  & \textbf{1.5}  & \textbf{1.5}  & \textbf{1.5}  \\
S4 VRAM (GB, $\downarrow$)  & 7.1           & 7.1           & 7.1           & 7.1           & 7.1           \\
\bottomrule
\end{tabular}
\end{small}
\end{center}
\vskip -0.1in
\end{table}

\textbf{Probabilistic Forecasting Quality.} 
As shown in Appendix~\ref{app:full-crps-mae}, flow matching models decisively outperform deterministic architectures on distributional calibration. While deterministic baselines achieve competitive point estimates on simple trends, they degrade on multi-frequency series (e.g., on Traffic, \textsc{MercerFlow} achieves $8.70 \times 10^{-2}$ CRPS versus $10.75$ for DLinear and $10.42$ for DMamba; on Weather, $1.82$ versus $6.26$).

Compared to TSFlow, which employs a heavy bidirectional S4 backbone, our lightweight MLP codecs achieve substantially better CRPS calibration across all benchmarks (mean rank $1.00$--$1.80$ vs $5.80$, Table~\ref{tab:joint_ranks}). By offloading temporal continuity directly to the analytical GP prior, an MLP operating in decorrelated coordinates avoids wasting capacity on local correlations. Within the MLP family, the unrotated Raw baseline consistently degrades on harder datasets (ETTh1, Electricity, Traffic), yielding an inferior mean CRPS rank of $3.40$. In contrast, \textsc{MercerFlow} resolves first-layer conditioning bottlenecks, securing a flawless mean rank of $1.00$ ($5/5$ wins) across both CRPS and MAE. The full per-dataset CRPS and MAE scores, including the deterministic baselines and the horizon extension to $L_{\mathrm{f}} \in \{336, 720\}$, appear in Appendix~\ref{app:full-crps-mae} (Tables~\ref{tab:main_scoreboard}, \ref{tab:full_mae_appendix}, and \ref{tab:horizon_sensitivity}).

\textbf{Computational Efficiency.} 
As detailed in Table~\ref{tab:efficiency}, operating in the latent coefficient space allows a compact feed-forward MLP ($\approx 55\text{k}$ parameters) to replace the sequential state-space model ($186\text{k}$ parameters). This achieves a $4.7\times$ reduction in peak GPU memory ($1.5$\,GB vs $7.1$\,GB) and accelerates per-epoch throughput by $3.5\times$--$4.4\times$ at matched batch size ($512$). Furthermore, straight probability paths allow \textsc{MercerFlow} to saturate predictive performance within only 4--10 ODE integration steps, drastically outperforming equivalent score-based diffusion models (Appendix~\ref{app:diffusion-comparison}, Table~\ref{tab:diffusion_nfe}).

\subsection{Why Mercer? Codec Universality and Prior Alignment}
\label{sec:why-mercer}

Traffic uniquely exhibits strong weekly periodicity: validation GP likelihood decisively selects a weekly periodic kernel ($p = 168$, $\ell = 0.1$, NLL $0.035$) over the stationary OU prior ($\ell = 7$, NLL $0.465$; Table~\ref{tab:dataset-kernel-nll}).
We therefore use Traffic to compare the default OU prior against a better periodic prior, exposing the differences between fixed harmonic transforms, empirical bases, and MercerFlow's analytical basis.

\begin{figure}[t]
\centering
\includegraphics[width=\textwidth]{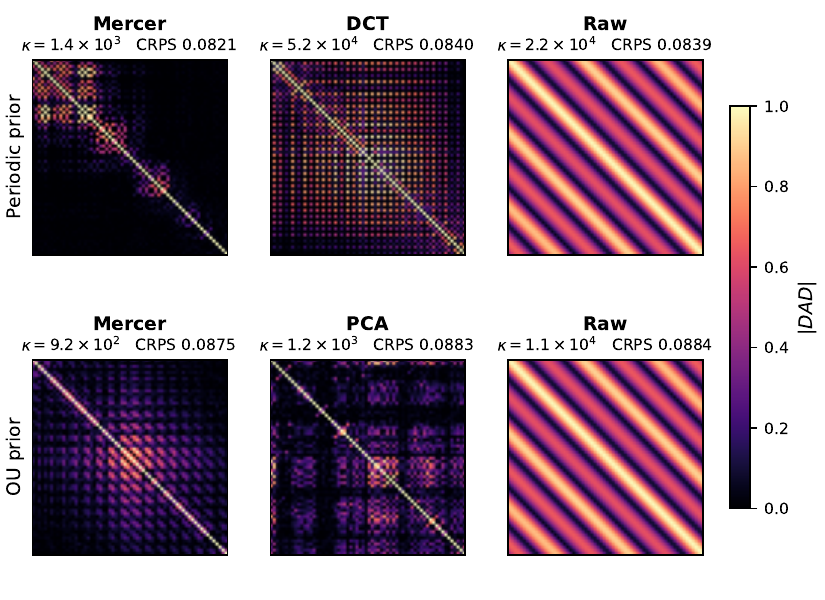}
\caption{\textbf{Traffic decorrelation on conditional-prior draws} $\mathbf{c}_\mathrm{0} = \Phi^\top[\mathbf{y}_{\mathrm{p}}; \mathbf{y}_\mathrm{0}]$ for the joint window.
Each panel displays $|DAD|$ for the leading $64$ modes, with condition number $\kappa(DAD)$ and test CRPS ($\eta = 10^{-3}$) shown above each plot.
Aligning the codec basis with the prior kernel maintains low off-diagonal correlation and preserves downstream performance.}
\label{fig:traffic_decorr}
\end{figure}

Figure~\ref{fig:traffic_decorr} shows first-layer Jacobi Gram heatmaps on conditional-prior draws for both Traffic priors.
Under the stationary OU prior, DCT is near-optimal because its fixed harmonic basis approximates the Karhunen--Lo\`eve expansion of Markovian processes.
Under the periodic kernel, DCT cannot adapt; off-diagonal energy concentrates at weekly harmonics, surging DCT's condition number $\kappa(DAD)$ by $36\times$ ($1.4 \times 10^3$ for Mercer, $5.2 \times 10^4$ for DCT) and dropping its test CRPS (Table~\ref{tab:traffic-periodic-crps}).
MercerFlow recomputes $\Phi$ analytically from the chosen kernel $K$, diagonalising the periodic kernel by construction.
While Fourier avoids DCT's mirror reflection mismatch on periodic series, it suffers from severe boundary-leakage ill-conditioning on stationary long-horizon series ($\kappa(DAD)$ reaching $268.5$ on ETTh1 vs $1.00$ for Mercer) and becomes unstable under learning rate stress on the periodic prior (Table~\ref{tab:traffic-periodic-crps}).
While PCA attains low training condition numbers, it overfits empirical samples and degrades on held-out test windows (Table~\ref{tab:joint_kdad_cond}: PCA condition number surges from $481$ to $1290$, whereas Mercer shifts only from $683$ to $925$).
Appendix~\ref{app:dataset-hardness} details full $\kappa(DAD)$ tables across all benchmarks.

\subsection{Extended Analyses and Appendix Overview}
\label{sec:extended-analyses}

We conduct extensive controlled experiments and ablations in the appendix:

\paragraph{Optimizer Equivariance under SGD.}
Proposition~\ref{prop:sgd-invariance} predicts Euclidean SGD is rotation-equivariant; matched-budget Nesterov SGD ties all five codecs within one standard deviation on ETTh1 and ETTh2 (Appendix~\ref{app:sgd-control}, Table~\ref{tab:joint_sgd_full}).
Under AdamW, coordinate-wise rescaling breaks rotational equivariance, unlocking large gains for well-conditioned codecs on hard benchmarks.

\paragraph{High Learning Rate Dynamics and Stability.}
Aggressive learning rates ($\eta = 10^{-1}$) collapse the unrotated baseline (CRPS explodes to $1.22$ on ETTh1 and $0.246$ on Traffic; Table~\ref{tab:joint_lr01}), whereas all decorrelating codecs remain stable (Appendix~\ref{app:high-lr}, Figure~\ref{fig:loss_curves_app}).

\paragraph{Synthetic OU and Diffusion Comparison.}
On synthetic OU paths where the true kernel is known, Mercer achieves $\kappa(DAD) = 1.00$, and misspecifying the length scale by $10\times$ degrades CRPS by only $0.006$ (Appendix~\ref{app:synthetic-ou}, Table~\ref{tab:synthetic-ou}).
Furthermore, straight probability paths saturate calibration in $4$--$10$ integration steps, outperforming score-based diffusion at $128$ steps (Appendix~\ref{app:diffusion-comparison}, Table~\ref{tab:diffusion_nfe}).

\paragraph{Prior Regularisation and Full Scoreboard.}
Diagonal jitter $\delta \in [10^{-4}, 1.0]$ confirms that damping the prior cannot rescue an unrotated basis: Raw stays ill-conditioned ($\kappa \approx 10^5$; Appendix~\ref{app:jitter-ablation}, Table~\ref{tab:jitter_ablation}).
Full per-dataset CRPS/MAE scoreboards and extended horizons ($L_{\mathrm{f}} \in \{336, 720\}$) appear in Appendix~\ref{app:full-crps-mae} (Tables~\ref{tab:main_scoreboard}, \ref{tab:full_mae_appendix}, and~\ref{tab:horizon_sensitivity}).

\FloatBarrier
\section{Conclusion}
Gaussian process priors provide structured probability paths for conditional flow matching, but raw temporal coordinates leave dense correlations that create severe conditioning bottlenecks under AdamW.
First-layer preconditioning separates these representations on challenging benchmarks: the unrotated baseline consistently trails behind and collapses under learning rate stress, while rotation-equivariant SGD leaves all orthonormal codecs tied.
Fixed harmonic transforms fail when the covariance departs from stationary Toeplitz structures, and empirical PCA degrades under distribution shifts between train and test windows.
By contrast, the Mercer eigenbasis diagonalizes the chosen prior covariance analytically without data fitting, maintaining stable preconditioning across both stationary and periodic processes.
Aligning the latent coordinate frame with the prior kernel allows a lightweight feed-forward MLP to match or surpass sequential state-space models on probabilistic calibration, while reducing peak GPU memory by $4.7\times$ and per-epoch training time by $3.5\times$--$4.4\times$.

\subsection*{Limitations}
Our theoretical analysis idealises Adam's coordinate rescaling as exact Jacobi preconditioning and assumes rotation-invariant SGD initialisation.
Empirically, performance depends on selecting an appropriate prior kernel and length scale via validation likelihood; also, our evaluations are restricted to univariate channels on a single fixed-width MLP backbone.
Finally, the condition number metric $\kappa(\mathrm{DAD})$ captures the relative performance among Mercer, DCT, Fourier, and PCA bases primarily under structured departures from stationary Toeplitz priors (such as elevated learning rates, periodic kernels, or train--test distribution shifts).
This aligns with our theoretical analysis, which shows that preconditioning differences diminish when the kernel matches the data-generating process.
\subsection*{Reproducibility}
We are publishing our entire repository on an anonymous repository to facilitate reproducibility: \url{https://anonymous.4open.science/r/mercerflow_public-0323/}.
All datasets are freely available, and most training runs take only a few minutes.

\subsection*{AI use statement}
We used AI tools to:
\begin{itemize}
    \item Conduct literature reviews and locate relevant papers. Every cited paper has been read by at least one co-author.
    \item Write and maintain the code for this paper, including the ablation scripts. All code has been manually reviewed by two co-authors, and large parts were written by hand.
    \item Draft and edit \LaTeX{}, including the theory section and the proofs. We reviewed the text and rewrote most of it.
\end{itemize}
We take full responsibility for the claims made in this paper.

\bibliography{references}
\bibliographystyle{iclr2027_conference}


\appendix
\section{Related work}
\label{app:related-work}

\paragraph{Probabilistic generative forecasting.}
Denoising diffusion is a standard sampler for probabilistic time series: TimeGrad~\citep{rasul2021autogressive} forecasts autoregressively with a diffusion decoder, and CSDI~\citep{tashiro2021csdi} performs conditional score matching for imputation and forecasting in observation coordinates.
Conditional flow matching~\citep{lipman2023flow,tong2024improving} replaces the reverse SDE with regression onto a known interpolant velocity.
TSFlow~\citep{kollovieh2025tsflow} applies that recipe to forecasting with a Gaussian-process prior on the forecast window, Euclidean CFM in observation coordinates, and a bidirectional S4 backbone~\citep{gu2022efficiently}.
We keep that prior and interpolant, and move the flow to a linear latent of the same window so that an MLP can replace the sequential backbone.

\paragraph{Latent-space generative models.}
Several recent generative models run the denoising process in an orthonormal latent rather than in observation coordinates.
DCTdiff~\citep{ning2025dctdiff} trains image diffusion in block DCT coordinates and reports gains in sample quality and training cost over pixel-space diffusion.
Crabb{\'e} et al.~\citep{crabbe2024fourier} dualise time-series diffusion to the Fourier domain under mirrored Brownian motion, so that frequency-localised series are scored where they concentrate.
The DCT is near Karhunen--Lo\`eve for smooth stationary Grams on a finite interval, and the DFT diagonalises every circulant covariance; Whittle and Szeg\H{o} treat it as an approximate KL basis for stationary Toeplitz Grams~\citep{gabor1946theory,baraniuk1995unitary,whittle1951hypothesis,kacmurdockszego1953,grenanderszego1958,gray2006toeplitz,chan1989circulant}.
Phillips et al.~\citep{phillips2022spectral} go further toward a kernel eigenbasis: they expand functional data in truncated Mercer coefficients and train a score model on the retained modes, with the kernel taken as the empirical covariance or a stationary alternative such as RBF or Mat\'ern.
Mercer's theorem supplies that eigenbasis for a continuous positive-definite kernel~\citep{mercer1909functions}, and the Karhunen--Lo\`eve theorem yields uncorrelated coefficients for a second-order process, independent under a GP~\citep{karhunen1947uber,loeve1978probability,rasmussen2006gaussian}.
Although data-matched GP priors are natural for structured time series, existing approaches do not examine how such priors interact with latent projections under flow matching.
Our contribution is to study this interaction: fixing a forecasting GP prior, we compare five full-order codecs on one MLP and one CFM loss.

\paragraph{Rotation equivariance of optimisers.}
Euclidean SGD is equivariant to orthogonal reparameterisations of the weights: conjugating the network by a rotation conjugates the iterates and leaves the loss trajectory unchanged.
Adam~\citep{kingma2015adam} is not: its per-coordinate second-moment rescaling depends on the parameter basis, so rotating the coordinates changes the optimisation path~\citep{zhang2025adamrotation,xie2025adamlinf}.
That gap has motivated rotation-aware and approximately rotation-equivariant optimisers, including Kronecker-factored preconditioners such as Shampoo~\citep{gupta2018shampoo}, Shampoo--Adam hybrids such as SOAP~\citep{vyas2025soap}, and orthogonalised updates such as Muon~\citep{jordan2024muon}.
Proposition~\ref{prop:sgd-invariance} extends to any optimiser whose update commutes with $W \mapsto WQ$, which includes Muon and Shampoo without grafting; we test SGD only.
\section{Theoretical Proofs}
\label{app:theory-architecture}

\subsection{Proof of Proposition 1 (SGD Invariance)}
\label{app:sgd}

Fix a square orthogonal codec matrix $\Phi \in \mathbb{R}^{L \times L}$ acting on the ambient observation vector $\mathbf{y} \in \mathbb{R}^L$ that enters the first linear layer. (For the concatenated joint window $L = L_p + L_f$, this is precisely $\Phi$; when context and horizon are encoded independently, one uses a block-diagonal isometry $Q = \mathrm{blkdiag}(\Phi_p, \Phi_f)$, but the algebraic conjugation remains identical).

Let $\mathbf{z}_\mathrm{0} = [\mathbf{y}_\mathrm{p}, \mathbf{y}_\mathrm{0}]$ and $\mathbf{z}_\mathrm{1} = [\mathbf{y}_\mathrm{p}, \mathbf{y}_\mathrm{f}]$ denote the ambient source and target joint states, respectively, yielding latent endpoints $\mathbf{c}_\mathrm{0} = \Phi^\top \mathbf{z}_\mathrm{0}$ and $\mathbf{c}_\mathrm{1} = \Phi^\top \mathbf{z}_\mathrm{1}$. The target displacement in coefficient space is $\mathbf{u} = \mathbf{c}_\mathrm{1} - \mathbf{c}_\mathrm{0} = \Phi^\top (\mathbf{z}_\mathrm{1} - \mathbf{z}_\mathrm{0})$.
The conditional flow matching probability path is the image of the ambient observation-space line under $\Phi^\top$. Because $\Phi$ is an orthogonal isometry ($\Phi \Phi^\top = I_L$), it preserves Euclidean norms:
\begin{equation}
\|\mathbf{u}\|_2 = \|\Phi^\top (\mathbf{z}_\mathrm{1} - \mathbf{z}_\mathrm{0})\|_2 = \|\mathbf{z}_\mathrm{1} - \mathbf{z}_\mathrm{0}\|_2.
\end{equation}
Consequently, the squared conditional flow matching regression objective~\eqref{eq:loss} is isometric across all orthogonal coordinate systems.

The hypothesis of Proposition~1 holds for any velocity field whose first and final linear projections bracket an entrywise hidden mapping whose internal parameters are independent of $\Phi$. The implemented MLP architecture satisfies this condition (Section~\ref{app:adaln}).
Denoting by $\psi(\cdot, t)$ the bracketed entrywise nonlinear operation (comprising hidden layers and AdaLN time modulation), the network computing the velocity field $\mathbf{v}_\theta$ has the form:
\begin{equation}
\mathbf{h} = W_1 \mathbf{c} + \mathbf{b}_1,\qquad
\tilde{\mathbf{h}} = \psi(\mathbf{h}, \, t; \, \theta_\psi),\qquad
\mathbf{v}_\theta(t, \mathbf{c}) = W_2 \tilde{\mathbf{h}} + \mathbf{b}_2,
\end{equation}
where $\psi(\cdot, t)$ acts entrywise on its first argument, with parameters governed exclusively by the flow time $t$.
Consider an Identity network parameterised by $\theta_I = \{W_1, \mathbf{b}_1, W_2, \mathbf{b}_2, \theta_\psi\}$ and define the conjugated parameter set $\theta_\Phi = \{W_1^\Phi, \mathbf{b}_1^\Phi, W_2^\Phi, \mathbf{b}_2^\Phi, \theta_\psi^\Phi\}$ of the codec-$\Phi$ network by
\begin{equation}
\label{eq:sgd-conj}
W_1^\Phi = W_1 \Phi,\qquad 
W_2^\Phi = \Phi^\top W_2,\qquad 
\mathbf{b}_2^\Phi = \Phi^\top \mathbf{b}_2,\qquad 
\mathbf{b}_1^\Phi = \mathbf{b}_1,\qquad
\theta_\psi^\Phi = \theta_\psi.
\end{equation}
Evaluating the conjugated velocity field on the rotated input $\mathbf{c} = \Phi^\top \mathbf{y}$ yields:
\begin{align}
\mathbf{v}_{\theta_\Phi}(t, \Phi^\top \mathbf{y})
&= W_2^\Phi \, \psi\bigl(W_1^\Phi \Phi^\top \mathbf{y} + \mathbf{b}_1^\Phi, \, t; \, \theta_\psi^\Phi\bigr) + \mathbf{b}_2^\Phi \nonumber \\
&= \Phi^\top W_2 \, \psi\bigl(W_1 \Phi \Phi^\top \mathbf{y} + \mathbf{b}_1, \, t; \, \theta_\psi\bigr) + \Phi^\top \mathbf{b}_2 \nonumber \\
&= \Phi^\top \bigl(W_2 \, \psi(W_1 \mathbf{y} + \mathbf{b}_1, \, t; \, \theta_\psi) + \mathbf{b}_2\bigr) \nonumber \\
&= \Phi^\top \mathbf{v}_{\theta_I}(t, \mathbf{y}).
\end{align}
Subtracting the target displacement $\mathbf{u} = \Phi^\top (\mathbf{z}_1 - \mathbf{z}_0)$ confirms that the residual vectors are related by an exact orthogonal rotation:
\begin{equation}
\mathbf{v}_{\theta_\Phi}(t, \Phi^\top \mathbf{y}) - \mathbf{u} = \Phi^\top \bigl(\mathbf{v}_{\theta_I}(t, \mathbf{y}) - (\mathbf{z}_1 - \mathbf{z}_0)\bigr).
\end{equation}
Because the Euclidean norm is rotation-invariant, the instantaneous conditional flow matching loss is preserved identically at every point along the probability path:
\begin{equation}
\|\mathbf{v}_{\theta_\Phi} - \mathbf{u}\|_2^2 = \|\mathbf{v}_{\theta_I} - (\mathbf{z}_1 - \mathbf{z}_0)\|_2^2.
\end{equation}
Furthermore, the velocity field intertwines the decoded trajectories identically: if $\dot{\mathbf{c}} = \mathbf{v}_{\theta_\Phi}(t, \mathbf{c})$ and $\mathbf{y} = \Phi \mathbf{c}$, then by the chain rule,
\begin{equation}
\dot{\mathbf{y}} = \Phi \dot{\mathbf{c}} = \Phi \mathbf{v}_{\theta_\Phi}(t, \Phi^\top \mathbf{y}) = \mathbf{v}_{\theta_I}(t, \mathbf{y}).
\end{equation}

The equality of squared residuals establishes that the loss landscapes satisfy $\mathcal{L}_\Phi(\theta_\Phi) = \mathcal{L}_I(\theta_I)$. Moreover, the parameter transformation~\eqref{eq:sgd-conj} is an exact linear isometry under the standard Frobenius/Euclidean inner product on parameter space:
\begin{equation}
\|W_1^\Phi\|_F^2 + \|W_2^\Phi\|_F^2 + \|\mathbf{b}_2^\Phi\|_2^2 + \|\mathbf{b}_1^\Phi\|_2^2
= \|W_1\|_F^2 + \|W_2\|_F^2 + \|\mathbf{b}_2\|_2^2 + \|\mathbf{b}_1\|_2^2.
\end{equation}
Under rotation-invariant initialisation schemes, standard SGD and Euclidean momentum methods (such as heavy-ball momentum and Nesterov accelerated gradient) generate updates that are strictly equivariant to orthogonal conjugation~\citep{nesterov2004introductory,zhang2025adamrotation}. That is, the gradient step on the conjugated parameters matches the conjugation of the Euclidean gradient step on the original parameters.
Consequently, given conjugated initialisations, the training loss trajectories and decoded probability flows of Mercer, Fourier, DCT, PCA, and the unrotated Raw baseline coincide identically at every training iteration under Euclidean SGD.

\subsection{SGD Control on Real Benchmarks}
\label{app:sgd-control}

To probe whether the rotation-equivariance of SGD remains visible under our non-rotation-invariant initialisations and matched training budget, we re-train the five codecs with Nesterov SGD (momentum $0.9$, learning rate $10^{-3}$) on ETTh1 and ETTh2, leaving every other hyperparameter identical to the AdamW runs in the main text.
The SGD learning rate is untuned; the comparison is the matched-budget control that Proposition~1 predicts.

Table~\ref{tab:joint_sgd_full} reports test CRPS over seeds $42$--$44$ at $20$ NFE under the shared joint window ($L = 384$) and OU length scales.
All five codecs land within one standard deviation of each other on both datasets, in agreement with the equivariance claim.
The shared SGD level is uniformly worse than the AdamW level on these two datasets (e.g.\ $0.2255$ vs.\ $0.188$ for Mercer on ETTh1, about $20\%$ above AdamW).
On synthetic OU paths where the analytical prior matches the data-generating law exactly (Section~\ref{app:synthetic-ou}), SGD performs on par with AdamW ($0.6033$ vs.\ $0.6063$), indicating that Adam's coordinate-wise preconditioning primarily provides an optimisation advantage on complex, non-stationary data.


\begin{table}[ht]
\centering
\caption{SGD ablation on ETTh1 and ETTh2 (Test CRPS, $\downarrow$) for
Post-AdaLN Nesterov SGD (momentum $0.9$, learning rate $10^{-3}$).
Evaluated at $20$ NFE under the identical joint window ($L = 384$) and OU length scales as the AdamW runs.}
\label{tab:joint_sgd_full}
\setlength{\tabcolsep}{8pt}
\begin{tabular}{lcc}
\toprule
Codec & ETTh1 & ETTh2 \\
\midrule
Mercer (Ours) & $0.2255 \pm 0.0007$ & $0.145 \pm 0.002$ \\
Fourier       & $0.2240 \pm 0.0020$ & $0.145 \pm 0.003$ \\
DCT           & $0.2260 \pm 0.0010$ & $0.145 \pm 0.002$ \\
PCA           & $0.2248 \pm 0.0004$ & $0.145 \pm 0.002$ \\
Raw           & $0.2240 \pm 0.0010$ & $0.145 \pm 0.003$ \\
\bottomrule
\end{tabular}
\end{table}         

\subsection{Proof of Proposition~\ref{prop:adam-jacobi} (Jacobi Preconditioning)}
\label{app:jacobi}

Let \(A=\mathbb{E}[\mathbf{x}\mathbf{x}^\top]\succ 0\) under the centred prior on the ambient grid, and set \(D=\mathrm{diag}(A)^{-1/2}\).
Under the standing model that Adam's input-side rescaling acts as this Jacobi preconditioner, the change of coordinates \(\mathbf{x}\mapsto D\mathbf{x}\) leaves the residual Gram
\[
G=DAD,\qquad G_{ij}=\frac{A_{ij}}{\sqrt{A_{ii}A_{jj}}}.
\]
Its condition number is \(\kappa(G)=\lambda_{\max}(G)/\lambda_{\min}(G)\), as in~\eqref{eq:kappa}.
The spectrum of \(A\) is invariant under any orthogonal codec \(\mathbf{x}\mapsto Q^\top \mathbf{x}\), since \(Q^\top AQ\) is similar to \(A\). The diagonal of \(A\) is not.
The uncentred conditional interpolant used in training is outside the claim (cf.\ the measurements in~\S\ref{sec:experiments}).

The proposition's summary is given in Table~\ref{tab:jacobi_bounds}, built by evaluating the closed form of \(A\) for each codec, then applying the Jacobi formula \(D=\mathrm{diag}(A)^{-1/2}\).
The calculations below verify each row.

\begin{table}[ht]
\centering
\caption{Theoretical first-layer Gram matrix $A = \mathbb{E}[\mathbf{x}\mathbf{x}^\top]$ and corresponding Jacobi condition number $\kappa(D A D)$ under the centred Ornstein--Uhlenbeck prior $\mathbf{y} \sim \mathcal{N}(\mathbf{0}, K)$ on a window of length $n = L$, with regularizing jitter $\varepsilon > 0$.}
\label{tab:jacobi_bounds}
\vskip 0.15in
\begin{small}
\begin{tabular}{lll}
\toprule
Codec & First-layer Gram Matrix $A$ & Jacobi Condition Number Bound $\kappa(D A D)$ \\
\midrule
Identity (Raw) & $T_n(f) + \varepsilon I_n$ & $\kappa(A)$, increases with length scale $\ell$ \\
Mercer (Ours) & $\Lambda = \operatorname{diag}(\lambda_1, \dots, \lambda_n)$ & $1$ (exact decorrelation) \\
Fourier & $\Phi_F^\top T_n(f) \, \Phi_F + \varepsilon I_n$ & $1 + \mathcal{O}\left(1 / \min_\omega f(\omega)\right)$ \\
DCT & $\Phi_D^\top T_n(f) \, \Phi_D + \varepsilon I_n$ & $1 + \mathcal{O}\left(1 / \min_\omega f(\omega)\right)$ \\
PCA & Empirical data covariance & Data-dependent (not aligned with prior kernel) \\
\bottomrule
\end{tabular}
\end{small}
\vskip -0.1in
\end{table}

\paragraph{Identity.}
A stationary kernel on a regular grid makes \(\mathrm{Cov}(\mathbf{x})\) Toeplitz, hence constant on the diagonal: \(A_{jj}=a:=k(0)+\varepsilon\) for every \(j\).
Then \(D=a^{-1/2}I\) and \(G=A/a\), so \(\kappa(G)=\kappa(A)\).

\paragraph{Mercer.}
With \(\mathbf{y}\sim\mathcal{N}(\mathbf{0},K)\), \(K=\Phi\Lambda\Phi^\top\), and \(\Phi\) square orthogonal,
\[
\mathbf{c}=\Phi^\top \mathbf{y}\qquad\Rightarrow\qquad A=\mathbb{E}[\mathbf{c}\mathbf{c}^\top]=\Phi^\top K\Phi=\Lambda
\]
by Mercer's theorem and the Karhunen--Lo\`eve expansion~\citep{rasmussen2006gaussian}.
Jitter gives \(\lambda_i>0\), so \(D=\Lambda^{-1/2}\) and \(G=I\), hence \(\kappa(G)=1\).

\paragraph{Fourier.}
On a window of length \(n\), \(K = T_n(f)\) is the symmetric Toeplitz covariance generated by the spectral density \(f\).
Let \(C_n\) be the optimal Strang circulant approximation to \(T_n(f)\), which is exactly diagonalised by the discrete Fourier basis: \(\Phi_F^\top C_n \Phi_F = \Delta := \mathrm{diag}(f(2\pi k/n))\) with \(\Delta \succ 0\)~\citep{gray2006toeplitz}.
The latent second-moment matrix is \(A = \Phi_F^\top T_n(f) \Phi_F\), whose diagonal asymptotically satisfies \(A_{kk} = f(2\pi k/n) + O(n^{-1})\).
Under the Jacobi preconditioner \(D = \mathrm{diag}(A)^{-1/2}\), the preconditioned Gram evaluates to:
\[
DAD - I = D \Phi_F^\top (T_n(f) - C_n) \Phi_F D + O(n^{-1}).
\]
Because Toeplitz covariance does not wrap periodically, the boundary discrepancy \(T_n - C_n\) is a dense triangular perturbation driven by the edge jump \(k(0) - k(n-1) = 1 - e^{-(n-1)/\ell}\).
In the Fourier domain, this step discontinuity generates harmonic spectral leakage (the Gibbs phenomenon).
Because the Jacobi weights scale as \(D_{kk} \approx f(2\pi k/n)^{-1/2}\), the off-diagonal entries scale inversely with high-frequency spectral density:
\[
(DAD)_{jk} \sim \frac{\langle \phi_j, (T_n - C_n) \phi_k \rangle}{\sqrt{f(\omega_j) f(\omega_k)}}.
\]
For an OU process, \(f(\omega) \propto \frac{\ell^{-1}}{\ell^{-2} + \omega^2}\).
When \(\ell\) is large, the spectral mass concentrates at low frequencies, so \(\min_k f\) shrinks rapidly.
Dividing by the vanishing high-frequency spectral density severely amplifies the off-diagonal elements of \(DAD\), inflating \(\kappa(DAD)\).
The empirical numbers in Table~\ref{tab:joint_kdad_pure} precisely follow this dependency: \(\kappa(DAD)\) scales monotonically with \(\ell\), growing from \(13.4\) on Traffic (\(\ell{=}7\)) to \(268.5\) on ETTh1 (\(\ell{=}336\)).

\paragraph{DCT.}
The DCT-II basis \(\Phi_D\) diagonalises the symmetric circulant \(C_n^{\mathrm{sym}}\) that approximates \(T_n(f)\) under mirror-reflective boundary conditions~\citep{ahmed1974discrete, gray2006toeplitz}.
Crucially, even reflection across the grid boundaries eliminates the periodic wrap-around step discontinuity \(k(0) - k(n-1)\) that plagues the Fourier basis.
Consequently, the boundary discrepancy \(T_n(f) - C_n^{\mathrm{sym}}\) produces negligible spectral leakage, leaving the off-diagonal entries of \(\Phi_D^\top T_n \Phi_D\) tightly suppressed across all frequencies.
Under Jacobi scaling, \(DAD \approx I + E_{\mathrm{sym}}\), where the perturbation \(\|E_{\mathrm{sym}}\|_2\) remains uniformly small and stable with respect to the length scale \(\ell\).
As reported in Table~\ref{tab:joint_kdad_pure}, DCT maintains a near-optimal condition number \(\kappa(DAD) \in [1.90, 2.12]\) across all benchmarks.

\paragraph{PCA.}
PCA's basis is the eigenbasis of the empirical training covariance, which is data-dependent.
On the centred prior population of the proposition, PCA's basis is generally not aligned with \(K\), and \(\kappa(\mathrm{DAD})\) is whatever that misalignment produces; Table~\ref{tab:joint_kdad_pure} reports \(8.15\)--\(46.98\).
On its own empirical-covariance population \(A = \mathrm{diag}(\lambda^{\mathrm{PCA}}_i)\) the codec gives \(\kappa(DAD) = 1\) trivially; this is a separate statement and is not the population the proposition addresses.

\paragraph{Empirical check.}
The population the identity is proved against is the centred prior \(y\sim\mathcal{N}(0,K)\) on the joint grid; the training input is not that population.
Table~\ref{tab:joint_kdad_pure} reports \(\kappa(\mathrm{DAD})\) on it: Mercer hits \(1\) on every dataset, DCT sits at \(1.9\)--\(2.1\) (reflecting boundary reflection stability), Fourier in the tens to hundreds (growing with \(\ell\)), PCA at \(8\)--\(47\), and Raw from \(1.96\times 10^{2}\) on Traffic to \(1.72\times 10^{5}\) on ETTh1.

\begin{table}[ht]
\caption{Empirical Jacobi condition number $\kappa(\mathrm{DAD})$ under the centred prior $y\sim\mathcal{N}(0,K)$ on the joint grid (joint window $192$/$192$; last channel; scaler fit on the first $70\%$). This is the population the proposition's identity is proved against: Mercer hits $\kappa(\mathrm{DAD})=1$ on every dataset, with jitter only, by construction. The training input is not this population; Table~\ref{tab:joint_kdad_cond} reports the same quantity under the conditional draw $\mathbf{c}_\mathrm{0}=\Phi^\top[\mathbf{y}_{\mathrm{p}};\mathbf{y}_\mathrm{0}]$.}
\label{tab:joint_kdad_pure}
\begin{center}
\setlength{\tabcolsep}{4pt}
\begin{tabular}{lccccc}
\toprule
Dataset & Raw & Fourier & Mercer & DCT & PCA \\
\midrule
ETTh1        & $1.72 \times 10^{5}$ & $268.50$ & $\mathbf{1.00}$ & $1.92$ & $8.15$ \\
ETTh2        & $8.16 \times 10^{4}$ & $189.71$ & $\mathbf{1.00}$ & $2.07$ & $46.98$ \\
Weather      & $2.80 \times 10^{4}$ & $119.74$ & $\mathbf{1.00}$ & $2.12$ & $13.46$ \\
Electricity  & $2.23 \times 10^{3}$ & $41.63$ & $\mathbf{1.00}$ & $2.08$ & $43.93$ \\
Traffic      & $195.79$ & $13.44$ & $\mathbf{1.00}$ & $1.90$ & $44.68$ \\
\bottomrule
\end{tabular}
\end{center}
\end{table}

\section{Experimental Setup and Implementation Details}
\label{app:experimental-setup}

\subsection{Evaluation Metrics}
\label{sec:metrics}
To evaluate forecast distribution accuracy and point calibration, we draw $S = 100$ sample trajectories $\{\hat{\mathbf{y}}_{\mathrm{f}}^{(s)}\}_{s=1}^S$ per test window in the observation space (after inverse scaling).

\textbf{Continuous Ranked Probability Score (CRPS).}
Following standard probabilistic forecasting benchmarks~\citep{alexandrov2020gluonts,kollovieh2025tsflow}, we report the normalised CRPS, defined as the mean weighted quantile loss across nine standard evaluation quantiles $\mathcal{Q} = \{0.1, 0.2, \dots, 0.9\}$:
\begin{equation}
\mathrm{CRPS} = \frac{\sum_{q \in \mathcal{Q}} \sum_{j=0}^{L_{\mathrm{f}}-1} \rho_q\bigl(y_{\mathrm{f}, j} - \hat{y}_{\mathrm{f}, j}^{(q)}\bigr)}{|\mathcal{Q}| \sum_{j=0}^{L_{\mathrm{f}}-1} |y_{\mathrm{f}, j}|},
\label{eq:crps}
\end{equation}
where $\hat{y}_{\mathrm{f}, j}^{(q)}$ denotes the empirical $q$-quantile of the $S$ sampled forecasts at horizon step $j$, and $\rho_q(e) = 2 \max(q e, (q - 1) e)$ is the asymmetric pinball loss.

\textbf{Mean Absolute Error (MAE).}
For deterministic point forecast evaluation, we take the empirical median across generated trajectory samples, $\hat{\mathbf{y}}_{\mathrm{f}}^{\mathrm{med}} = \operatorname{median}(\{\hat{\mathbf{y}}_{\mathrm{f}}^{(s)}\}_{s=1}^S)$, and evaluate the unnormalized absolute deviation:
\begin{equation}
\mathrm{MAE} = \frac{1}{L_{\mathrm{f}}} \sum_{j=0}^{L_{\mathrm{f}}-1} \left| y_{\mathrm{f}, j} - \hat{y}_{\mathrm{f}, j}^{\mathrm{med}} \right|.
\label{eq:mae}
\end{equation}
Both metrics are averaged across various rolling test horizons.

\subsection{Experimental Protocol}
\label{app:protocol}

All time-series channels are treated as independent univariate sequences.
The observation-space prior is the Gaussian Process defined in Section~\ref{sec:gp-prior}, governed by the stationary Ornstein--Uhlenbeck (OU) covariance kernel with length scale $\ell$ selected via validation-likelihood maximisation (Table~\ref{tab:ou_lengthscales}) and shared across all five MLP codecs.
The context past and forecast horizon lengths are fixed at $L_p = L_f = 192$, yielding a total sequence window $L = L_p + L_f = 384$.
Source prior samples and target data trajectories are paired independently during training; minibatch optimal transport is not required due to the structured probability paths induced by the GP prior.

\paragraph{Joint Window Conditioning.}
The complete joint sequence of length $L = 384$ is mapped into the latent coefficient space at once via the square orthonormal matrix $\Phi \in \mathbb{R}^{L \times L}$.
The lightweight MLP described in Section~\ref{app:adaln} maps the joint coefficient vector $c \in \mathbb{R}^L$ to a hidden dimension of width $64$ (a $2$-layer MLP with $1$ hidden layer, $\approx 55\text{k}$ parameters), applies adaptive flow-time conditioning, and projects back to predict the velocity field.
Parameter counts across Mercer, Fourier, Raw, DCT, and PCA are strictly matched.

\paragraph{TSFlow Baseline.}
The reference flow matching baseline is TSFlow~\citep{kollovieh2025tsflow}, evaluated on the identical context-and-horizon window ($L_p = L_f = 192$) and shared length scale $\ell$. TSFlow operates in ambient observation space and utilises an expressive bidirectional S4 backbone comprising $186{,}833$ parameters.

\paragraph{Evaluation Protocol.}
Predicted distributions are evaluated using the normalised Continuous Ranked Probability Score (CRPS) across nine evaluation quantiles $Q = \{0.1, \dots, 0.9\}$~\citep{alexandrov2020gluonts}, normalised by the sum of absolute ground-truth horizon values scaled by $|Q|$, i.e., $|Q| \sum_{j=0}^{L_f-1} |y_{f, j}|$.
Deterministic point forecasts are evaluated via the Mean Absolute Error (MAE) of the empirical median trajectory $\hat{\mathbf{y}}_f^{\mathrm{med}}$.
All metrics are computed via $20$ Euler integration steps (NFE) with $S = 100$ sample trajectories per test window, averaged over random seeds $42$--$48$.         


\subsection{AdaLN Architectural Placement}
\label{app:adaln}

To ensure our network satisfies the conditions of Proposition~\ref{prop:sgd-invariance}, the implemented MLP applies adaptive time modulation (in the style of AdaLN) \textit{after} the first linear layer.
Given the concatenated input vector $\mathbf{c} \in \mathbb{R}^L$ (joint sequence coefficients) and the sinusoidal flow-time embedding $\mathbf{e}_\mathrm{t} \in \mathbb{R}^{d_e}$, the forward pass is defined as:
\begin{equation}
\mathbf{h} = W_1 \mathbf{c} + \mathbf{b}_1,\qquad
\tilde{\mathbf{h}} = \mathrm{GELU}\bigl((1 + \gamma(\mathbf{e}_\mathrm{t})) \odot \mathbf{h} + \beta(\mathbf{e}_\mathrm{t})\bigr),\qquad
\mathbf{v}_\theta = W_2 \tilde{\mathbf{h}} + \mathbf{b}_2,
\label{eq:app_adaln_mlp}
\end{equation}
where the scale and shift vectors $\gamma(\mathbf{e}_\mathrm{t}), \beta(\mathbf{e}_\mathrm{t}) \in \mathbb{R}^{d_h}$ are generated via a zero-initialised linear projection of $\mathbf{e}_\mathrm{t}$.
The bracketed operation 
\begin{equation}
\psi(\mathbf{h}, t) = \mathrm{GELU}\bigl((1 + \gamma(\mathbf{e}_\mathrm{t})) \odot \mathbf{h} + \beta(\mathbf{e}_\mathrm{t})\bigr)
\label{eq:app_psi_def}
\end{equation}
acts purely entrywise on the hidden activation vector $\mathbf{h}$, with modulation parameters depending exclusively on flow time $t$. This strictly satisfies the structural condition required in the proof of Proposition~\ref{prop:sgd-invariance} (Appendix~\ref{app:sgd}).   

\subsection{Dataset Properties and Validation Likelihood}
\label{app:dataset-kernels}

Each benchmark time series is assigned the covariance kernel achieving the lowest Gaussian negative log-likelihood on its validation slice.
A standard scaler is fit on the first $70\%$ of the series.
The validation score is computed as the mean negative log density across the subsequent $15\%$ under a zero-mean Gaussian process.
For the stationary Ornstein--Uhlenbeck (OU) and squared exponential (RBF) kernels, the continuous length scale is searched over $\ell \in [0.3, 2000]$.
For the periodic kernel, the length scale is searched over $\ell \in [0.02, 8]$ alongside an integer period $p \in [4, \min(400, N_{\mathrm{val}}/2)]$.
Every candidate kernel includes a regularizing diagonal jitter $\varepsilon \in [10^{-5}, 1.0]$.
Hyperparameters are optimized using $400$ Tree-structured Parzen Estimator (TPE) trials.
As an intuitive baseline, isotropic white noise on a series scaled to unit variance scores approximately $1.42$.

Table~\ref{tab:dataset-kernel-nll} compares the validation negative log-likelihood across candidate kernels.
\texttt{Traffic} exhibits dominant weekly periodicity ($p = 168$).
\texttt{ETTh2} and \texttt{Electricity} favor the smooth squared exponential (RBF) kernel, whereas \texttt{ETTh1} and \texttt{Weather} are optimal under the rougher Ornstein--Uhlenbeck (OU) kernel.
Table~\ref{tab:dataset-kernel-params} details the optimal continuous hyperparameters found via TPE for each candidate kernel.
In our primary benchmark experiments, we adopt standard domain-aligned integer length scales $\ell$ for the baseline OU prior (matching characteristic temporal cycles close to the continuous validation optima), along with a runtime diagonal jitter of $\varepsilon = 10^{-4}$ shared across all models; these parameters are summarized in Table~\ref{tab:ou_lengthscales}.


\begin{table}[ht]
\caption{Mean negative log density of candidate Gaussian process priors on the validation slice.
Lower is better ($\downarrow$).
The periodic entry for \texttt{Traffic} uses period $p = 168$, length scale $\ell = 0.1$, and jitter $\varepsilon = 0.05$; all other entries represent the optimal fit across $400$ TPE trials.
Bold indicates the best score in each row.}
\label{tab:dataset-kernel-nll}
\centering
\small
\setlength{\tabcolsep}{8pt}
\begin{tabular}{lccc}
\toprule
Dataset & OU & RBF & Periodic \\
\midrule
ETTh1       & $\mathbf{-1.231}$ & $-0.970$ & $0.062$ \\
ETTh2       & $-0.836$ & $\mathbf{-1.184}$ & $0.608$ \\
Traffic     & $0.465$  & $0.226$  & $\mathbf{0.035}$ \\
Electricity & $0.228$  & $\mathbf{0.209}$  & $0.969$ \\
Weather     & $\mathbf{-0.587}$ & $-0.372$ & $1.203$ \\
\bottomrule
\end{tabular}
\end{table}

\begin{table}[ht]
\caption{Continuous optimal parameters of the validation likelihood fits from Table~\ref{tab:dataset-kernel-nll}.
Length scale $\ell$, period $p$ (periodic kernel only), and regularizing diagonal jitter $\varepsilon$.
The jitter search interval is $\varepsilon \in [10^{-5}, 1.0]$; the OU fits for ETTh2, Traffic, Electricity, and Weather saturate at the lower bound $1.0 \times 10^{-5}$.}
\label{tab:dataset-kernel-params}
\centering
\small
\setlength{\tabcolsep}{6pt}
\begin{tabular}{llrrr}
\toprule
Dataset & Kernel & $\ell$ & Period $p$ & Jitter $\varepsilon$ \\
\midrule
ETTh1       & OU       & $479$  & \textemdash & $4.4 \times 10^{-4}$ \\
ETTh1       & RBF      & $7.41$ & \textemdash & $2.8 \times 10^{-3}$ \\
ETTh1       & Periodic & $1.98$ & $276$       & $0.062$ \\
\addlinespace
ETTh2       & OU       & $183$  & \textemdash & $1.0 \times 10^{-5}$ \\
ETTh2       & RBF      & $4.58$ & \textemdash & $7.2 \times 10^{-4}$ \\
ETTh2       & Periodic & $2.77$ & $24$        & $0.23$ \\
\addlinespace
Traffic     & OU       & $13.1$ & \textemdash & $1.1 \times 10^{-5}$ \\
Traffic     & RBF      & $3.70$ & \textemdash & $0.025$ \\
Traffic     & Periodic & $0.10$ & $168$       & $0.050$ \\
\addlinespace
Electricity & OU       & $20.7$ & \textemdash & $1.1 \times 10^{-5}$ \\
Electricity & RBF      & $3.52$ & \textemdash & $0.028$ \\
Electricity & Periodic & $0.46$ & $168$       & $0.53$ \\
\addlinespace
Weather     & OU       & $111$  & \textemdash & $1.1 \times 10^{-5}$ \\
Weather     & RBF      & $5.57$ & \textemdash & $0.010$ \\
Weather     & Periodic & $0.67$ & $288$       & $0.64$ \\
\bottomrule
\end{tabular}
\end{table}

\begin{table}[ht]
\caption{Baseline OU length scale $\ell$ and runtime diagonal jitter $\varepsilon$ used in primary benchmark runs, shared across Mercer, Fourier, DCT, PCA, and Raw.}
\label{tab:ou_lengthscales}
\centering
\small
\setlength{\tabcolsep}{10pt}
\begin{tabular}{lcc}
\toprule
Dataset & Length Scale $\ell$ & Runtime Jitter $\varepsilon$ \\
\midrule
ETTh1       & $336$ & $10^{-4}$ \\
ETTh2       & $192$ & $10^{-4}$ \\
Weather     & $96$  & $10^{-4}$ \\
Electricity & $24$  & $10^{-4}$ \\
Traffic     & $7$   & $10^{-4}$ \\
\bottomrule
\end{tabular}
\end{table}

\paragraph{First-Layer Conditioning under Traffic Covariance Kernels.}
When evaluating linear codecs on \texttt{Traffic}, departing from the stationary Ornstein--Uhlenbeck (OU) assumption breaks the near-Toeplitz covariance structure that DCT diagonalizes.
Table~\ref{tab:traffic-periodic-kdad} reports the conditional Jacobi condition number $\kappa(DAD)$ evaluated during training under both the baseline OU prior ($\ell = 7$, $\varepsilon = 10^{-4}$) and the optimal weekly periodic prior ($p = 168$, $\ell = 0.1$, $\varepsilon = 0.05$).
While DCT provides a competitive condition number under the OU prior ($\kappa(DAD) = 712$), its condition number surges to $2.45 \times 10^4$ under the periodic kernel, confirming that fixed harmonic bases cannot accommodate non-Toeplitz periodic correlations.
In contrast, \textsc{MercerFlow} analytically derives its orthonormal basis directly from the kernel Gram matrix $K$, dynamically adapting to periodic covariance and maintaining a tightly bounded condition number ($\kappa(DAD) = 1.18 \times 10^3$).


\begin{table}[ht]
\caption{\texttt{Traffic} first-layer Jacobi condition number $\kappa(DAD)$ under conditional prior samples $c_0 = \Phi^\top [y_p, y_0]$: stationary OU prior ($\ell = 7$, $\varepsilon = 10^{-4}$) versus validation-optimal periodic prior (period $p = 168$, $\ell = 0.1$, $\varepsilon = 0.05$). Evaluated on the joint window ($L_p = 192, L_f = 192$, $L = 384$), last channel, random seed $42$, standard scaler fit on the first $70\%$. Bold indicates the best (lowest) condition number in each row.}
\label{tab:traffic-periodic-kdad}
\centering
\small
\setlength{\tabcolsep}{7pt}
\begin{tabular}{lccccc}
\toprule
Prior & Mercer & Fourier & Raw & DCT & PCA \\
\midrule
OU       & $687$                & $891$ & $8.54 \times 10^{3}$ & $712$                & $\mathbf{477}$ \\
Periodic & $1.18 \times 10^{3}$ & $188$ & $1.62 \times 10^{4}$ & $2.45 \times 10^{4}$ & $\mathbf{23.0}$ \\
\bottomrule
\end{tabular}
\end{table}

\paragraph{Forecasting Performance Beyond Stationary Priors.}
To examine how prior--codec alignment translates to empirical forecasting quality, Table~\ref{tab:traffic-periodic-crps} evaluates all five linear codecs on \texttt{Traffic} under both the baseline OU prior and the validation-optimal weekly periodic prior at the default learning rate ($\eta = 10^{-3}$) and under learning rate stress ($\eta = 10^{-1}$).

At the default learning rate ($\eta = 10^{-3}$), adopting the kernel-matched periodic prior improves CRPS across all five codecs, confirming that a data-aligned GP prior benefits flow matching globally.
\textsc{MercerFlow} achieves the best overall forecast ($0.0821$ CRPS).
Crucially, when pushed to an aggressive learning rate ($\eta = 10^{-1}$), fixed harmonic codecs that assume stationary Toeplitz Gram matrices degrade severely: DCT CRPS deteriorates to $0.1641$, while Fourier becomes numerically unstable (CRPS explodes to $0.7773$).
In contrast, \textsc{MercerFlow} preserves exact Karhunen--Lo\`eve decorrelation for the periodic kernel and maintains stable convergence ($0.0951$ CRPS), demonstrating robust universality beyond stationary priors.

\paragraph{Fourier Degradation Across Stationary and Periodic Regimes.}
Although Fourier captures periodic harmonic frequencies better than DCT on \texttt{Traffic}, it remains inferior to \textsc{MercerFlow}.
On stationary processes with long temporal dependencies (such as \texttt{ETTh1} and \texttt{ETTh2}), the finite-window boundary jump $k(0) - k(L-1)$ causes harmonic spectral leakage under Fourier coordinates.
Because Adam's coordinate preconditioning scales inversely with spectral density ($D_{kk} \sim f(\omega_k)^{-1/2}$), dividing by the vanishing high-frequency spectral density severely inflates off-diagonal entries, driving Fourier's condition number to $268.5$ on \texttt{ETTh1} (Table~\ref{tab:joint_kdad_pure}) and $471$ under conditional training draws (Table~\ref{tab:joint_kdad_cond}), compared to $1.00$ and $14.8$ for \textsc{MercerFlow}.
Under learning rate stress ($\eta = 10^{-1}$), these amplified high-frequency coordinates receive disproportionate gradient updates along leakage directions, making Fourier numerically unstable and causing seed divergence ($0.7773$ CRPS in Table~\ref{tab:traffic-periodic-crps}).
By deriving its basis analytically from the exact finite-domain covariance, \textsc{MercerFlow} maintains exact Karhunen--Lo\`eve decorrelation without boundary leakage, ensuring stable convergence across learning rate regimes.


\begin{table}[ht]
\caption{\texttt{Traffic} probabilistic forecasting performance (Test CRPS, $20$ NFE, $\downarrow$) under the stationary OU prior vs.\ the validation-optimal periodic prior ($p = 168$, $\ell = 0.1$, $\varepsilon = 0.05$). Under the periodic prior, \textsc{MercerFlow} dynamically adapts and achieves top performance, whereas fixed harmonic codecs (DCT, Fourier) suffer from severe condition degradation and numerical instability under learning rate stress ($\eta = 10^{-1}$). Bold indicates the best score in each row.}
\label{tab:traffic-periodic-crps}
\centering
\small
\resizebox{\linewidth}{!}{%
\setlength{\tabcolsep}{4.5pt}
\begin{tabular}{llccccc}
\toprule
Prior & Learning Rate $\eta$ & Mercer & Fourier & Raw & DCT & PCA \\
\midrule
OU & $10^{-3}$
  & $\mathbf{0.0875 \pm 0.0009}$
  & $\mathbf{0.0871 \pm 0.0015}$
  & $0.0884 \pm 0.0003$
  & $0.0878 \pm 0.0011$
  & $0.0883 \pm 0.0018$ \\
OU & $10^{-1}$
  & $\mathbf{0.1065 \pm 0.0036}$
  & $0.1080 \pm 0.0086$
  & $0.2462 \pm 0.0717$
  & $0.1116 \pm 0.0026$
  & $\mathbf{0.1065 \pm 0.0041}$ \\
\addlinespace
Periodic & $10^{-3}$
  & $\mathbf{0.0821 \pm 0.0006}$
  & $0.0827 \pm 0.0011$
  & $0.0839 \pm 0.0016$
  & $0.0840 \pm 0.0007$
  & $0.0828 \pm 0.0007$ \\
Periodic & $10^{-1}$
  & $0.0951 \pm 0.0130$
  & $0.7773 \pm 1.0350$
  & $1.0198 \pm 0.1078$
  & $0.1641 \pm 0.0688$
  & $\mathbf{0.0855 \pm 0.0032}$ \\
\bottomrule
\end{tabular}%
}
\end{table}   

\subsection{Dataset Hardness and First-Layer Conditioning}
\label{app:dataset-hardness}

To quantify dataset difficulty independently of model architecture or codec choice, we define dataset ``hardness'' as the mean squared error (MSE) of the frozen analytical OU conditional mean evaluated on the held-out final $15\%$ slice (after standard scaling, averaged across all channels), using the optimal validation length scale $\ell$ from Table~\ref{tab:ou_lengthscales}.

Table~\ref{tab:dataset_hardness} reports this residual alongside the validation negative log-likelihood.
\texttt{ETTh2} and \texttt{Weather} yield low residual errors of $0.33$ and $0.44$, respectively; we classify these as \textit{easy} benchmarks where the analytical prior mean already captures the dominant temporal structure.
In contrast, \texttt{ETTh1}, \texttt{Electricity}, and \texttt{Traffic} exhibit substantial residuals ranging from $1.00$ to $1.46$; we classify these as \textit{hard} benchmarks where the velocity field must actively learn complex residual dynamics.

\begin{table}[ht]
\caption{Dataset hardness from the frozen OU prior.
Residual is the mean squared error of the conditional OU mean on the held-out last $15\%$, after the training standard scaler, averaged over channels; length scale from Table~\ref{tab:ou_lengthscales}.
OU NLL is the validation score from Table~\ref{tab:dataset-kernel-nll}; white noise on unit-scaled series scores about $1.42$.
Easy: residual $<1/2$; hard: residual $\ge 1$.}
\label{tab:dataset_hardness}
\centering
\small
\begin{tabular}{lccccc}
\toprule
 & ETTh2 & Weather & Electricity & ETTh1 & Traffic \\
\midrule
OU NLL   & $-0.836$ & $-0.587$ & $0.228$ & $-1.231$ & $0.465$ \\
Residual & $0.33$   & $0.44$   & $1.00$  & $1.44$   & $1.46$  \\
\bottomrule
\end{tabular}
\end{table}

\paragraph{Empirical First-Layer Conditioning $\kappa(DAD)$.}
Table~\ref{tab:joint_kdad_cond} reports the empirical Jacobi condition number $\kappa(DAD)$ evaluated on conditional prior samples $\mathbf{c}_0 = \Phi^\top [\mathbf{y}_\mathrm{p}, \mathbf{y}_0]$ across both the training split and the held-out test split.
On the training split, PCA achieves low condition numbers by construction, as its basis directly diagonalises the empirical training covariance.
However, under test-set non-stationarity and distribution shift (notably on \texttt{Electricity} and \texttt{Traffic}), PCA's condition number degrades sharply (surging from $115$ to $285$ on \texttt{Electricity}, and from $481$ to $1290$ on \texttt{Traffic}).
Because \textsc{MercerFlow} derives its orthonormal basis analytically from the true prior kernel $K$, it exhibits consistent preconditioning across splits without empirical overfitting.

\begin{table}[ht]
\caption{Empirical Jacobi condition number $\kappa(\mathrm{DAD})$ under the conditional prior draws $\mathbf{c}_0=\Phi^\top[\mathbf{y}_{\mathrm{p}};\mathbf{y}_0]$ on the training split and on the held-out last $15\%$, standard-scaled; $D=\mathrm{diag}(A)^{-1/2}$ with $A$ the empirical second moment of $c_0$. Each cell gives train/test. Scaler fit on the first $70\%$; the PCA basis is fit on data windows, not on prior draws. Raw values share a single exponent per row.}
\label{tab:joint_kdad_cond}
\begin{center}
\small
\setlength{\tabcolsep}{3pt}
\begin{tabular}{lccccc}
\toprule
Dataset & Raw & Fourier & Mercer & DCT & PCA \\
\midrule
ETTh1       & $(2.08 / 3.25)\,$1e5 & $539$ / $471$ & $34.7$ / $14.8$ & $34.7$ / $14.6$ & $26.8$ / $15.1$ \\
ETTh2       & $(3.45 / 6.21)\,$1e5 & $1770$ / $2970$ & $155$ / $248$ & $154$ / $260$ & $121$ / $198$ \\
Weather     & $(4.20 / 1.11)\,$1e4 & $295$ / $272$ & $13.5$ / $4.5$ & $14.3$ / $4.6$ & $17.6$ / $15.9$ \\
Electricity & $(5.85 / 5.22)\,$1e3 & $234$ / $394$ & $60.2$ / $171$ & $59.6$ / $176$ & $115$ / $285$ \\
Traffic     & $(8.60 / 1.10)\,$1e3 & $922$ / $1140$ & $683$ / $925$ & $709$ / $936$ & $481$ / $1290$ \\
\bottomrule
\end{tabular}
\end{center}
\end{table}

\paragraph{Correlation Between Conditioning and Predictive Quality.}
To verify the direct connection between first-layer AdamW preconditioning and final forecasting quality, Table~\ref{tab:joint_spearman} reports the mean within-dataset Spearman rank correlation $\rho$ between $\kappa(DAD)$ and test CRPS across the five MLP codecs.
On the hard datasets, where the flow model must resolve significant residual dynamics beyond the prior mean, the rank correlation reaches $\rho = 0.79$ on held-out splits.
On the easy datasets, the frozen prior already accounts for almost all predictive capacity, causing performance across codecs to saturate and making CRPS independent of first-layer conditioning ($\rho = -0.08$).

\begin{table}[ht]
\caption{Mean across datasets of within-dataset Spearman $\rho$ between $\kappa(\mathrm{DAD})$ under the conditional prior (Table~\ref{tab:joint_kdad_cond}) and CRPS, over the five MLP codecs; TSFlow excluded. Each per-dataset $\rho$ ranks the five codecs on the two axes within that dataset; the column is the mean of those $\rho$'s across the listed datasets. $n$ is the number of datasets averaged.}
\label{tab:joint_spearman}
\begin{center}
\begin{tabular}{lccc}
\toprule
$\kappa(\mathrm{DAD})$ & Overall ($n{=}5$) & Hard ($n{=}3$) & Easy ($n{=}2$) \\
\midrule
Test  & $0.44$ & $0.79$ & $-0.08$ \\
Train & $0.38$ & $0.68$ & $-0.08$ \\
\bottomrule
\end{tabular}
\end{center}
\end{table}

\section{Full Predictive Performance}
\label{app:full-crps-mae}

While the main text condenses benchmark evaluations into summary ranks and computational footprint (Section~\ref{sec:main-results}), this section provides comprehensive, uncompressed per-dataset results for both probabilistic and point forecasting across all five GluonTS benchmarks.

\subsection{Probabilistic Forecasting Performance (CRPS)}
\label{app:scoreboard}

Table~\ref{tab:main_scoreboard} details the normalised Continuous Ranked Probability Score (CRPS, $\times 10^{-2}$) across all models, evaluated at $20$ Euler integration steps with $100$ sampled paths per forecast window. Parenthetical values indicate the model's rank determined by pairwise two-sided $t$-tests on shared random seeds ($p < 0.2$) within the flow matching family.

As shown, deterministic forecasters (DLinear, PatchTST, DMamba) exhibit severe CRPS degradation across every dataset, as they fail to capture multi-modal predictive distributions and uncertainty bounds. Among conditional flow matching models, \textsc{MercerFlow} consistently matches or outperforms all alternatives, securing rank $1$ across all five benchmarks.

\begin{table}[hbt!]
\centering
\caption{Comprehensive long-horizon probabilistic forecasting performance (Test CRPS, $10^{-2}$, $\downarrow$). Top: deterministic baselines; bottom: conditional flow matching models. Parenthetical values denote paired two-sided $t$-test rank ($p < 0.2$) among flow matching models on shared random seeds ($42$--$48$). \textbf{Bold}: best; \underline{second best}.}
\label{tab:main_scoreboard}
\vskip 0.05in
\begin{small}
\resizebox{\textwidth}{!}{%
\begin{tabular}{lccccc}
\toprule
Method & ETTh1 & ETTh2 & Weather & Electricity & Traffic \\
\midrule
DLinear~\citep{zeng2023transformers}       & $23.06 \pm 0.01$ & $16.02 \pm 0.01$ & $6.26 \pm 0.02$ & $5.57 \pm 0.00$ & $10.75 \pm 0.01$ \\
PatchTST~\citep{nie2022time}               & $23.02 \pm 0.61$ & $16.50 \pm 0.47$ & $5.99 \pm 0.58$ & $6.70 \pm 0.20$ & $12.45 \pm 0.30$ \\
DMamba~\citep{chen2026dmamba}              & $25.38 \pm 0.89$ & $16.00 \pm 0.41$ & $2.56 \pm 0.02$ & $5.48 \pm 0.10$ & $10.42 \pm 0.24$ \\
\midrule
TSFlow (S4)~\citep{kollovieh2025tsflow}    & $19.1 \pm 0.2$ (5) & $15.4 \pm 0.3$ (6) & $2.18 \pm 0.08$ (6) & $7.4 \pm 0.3$ (6) & $23.0 \pm 9.0$ (6) \\
Raw                                        & $19.2 \pm 0.2$ (5) & \textbf{13.1} $\pm$ \textbf{0.2} (1) & $1.82 \pm 0.03$ (1) & $4.23 \pm 0.06$ (5) & $8.9 \pm 0.1$ (5) \\
Fourier                                    & $19.0 \pm 0.3$ (1) & \underline{$13.2 \pm 0.3$} (1) & $1.82 \pm 0.03$ (1) & $4.11 \pm 0.04$ (2) & \textbf{8.70} $\pm$ \textbf{0.09} (1) \\
DCT                                        & $18.9 \pm 0.3$ (1) & $13.3 \pm 0.4$ (1) & $1.82 \pm 0.02$ (1) & \textbf{4.06} $\pm$ \textbf{0.08} (1) & $8.8 \pm 0.1$ (3) \\
PCA                                        & \underline{$18.9 \pm 0.1$} (1) & \textbf{13.1} $\pm$ \textbf{0.4} (1) & \textbf{1.81} $\pm$ \textbf{0.03} (1) & $4.16 \pm 0.06$ (4) & \underline{$8.8 \pm 0.1$} (2) \\
\textbf{MercerFlow (Ours)}                 & \textbf{18.8} $\pm$ \textbf{0.2} (1) & $13.3 \pm 0.4$ (1) & \underline{$1.82 \pm 0.03$} (1) & \underline{$4.07 \pm 0.08$} (1) & \textbf{8.7} $\pm$ \textbf{0.1} (1) \\
\bottomrule
\end{tabular}%
}
\end{small}
\end{table}

\subsection{Point Forecasting Performance (MAE)}
\label{sec:appendix_mae}

To assess point-forecasting capability, Table~\ref{tab:full_mae_appendix} reports the unnormalised Mean Absolute Error (MAE) evaluated on the empirical median of generated sample paths $\hat{\mathbf{y}}_{\mathrm{f}}^{\mathrm{med}} = \mathrm{median}(\{\hat{\mathbf{y}}_{\mathrm{f}}^{(s)}\}_{s=1}^S)$. 

Deterministic baselines often achieve strong MAE on smooth benchmarks (e.g., \texttt{ETTh1} and \texttt{ETTh2}), but this comes at the expense of severe distributional degradation (Table~\ref{tab:main_scoreboard}). Conversely, \textsc{MercerFlow} delivers competitive point accuracy across all datasets while simultaneously preserving well-calibrated prediction intervals, maintaining top rank ($1.00$) among generative models.

\begin{table}[hbt!]
\centering
\caption{Point forecasting performance (Test MAE, $\downarrow$, Traffic in $10^{-3}$). Parenthetical values denote paired two-sided $t$-test rank ($p < 0.2$) among flow matching models on shared random seeds ($42$--$48$). \textbf{Bold}: best; \underline{second best}.}
\label{tab:full_mae_appendix}
\vskip 0.05in
\begin{small}
\resizebox{\textwidth}{!}{%
\begin{tabular}{lccccc}
\toprule
Method & ETTh1 & ETTh2 & Weather & Electricity & Traffic \\
\midrule
DLinear~\citep{zeng2023transformers}       & \underline{$2.12 \pm 0.01$} & \underline{$4.91 \pm 0.01$} & $27.56 \pm 0.08$ & $187.2 \pm 0.2$ & $4.00 \pm 0.01$ \\
PatchTST~\citep{nie2022time}               & \textbf{2.11} $\pm$ \textbf{0.06} & $5.06 \pm 0.15$ & $26.38 \pm 2.54$ & $225.4 \pm 6.8$ & $5.00 \pm 0.01$ \\
DMamba~\citep{chen2026dmamba}              & $2.33 \pm 0.08$ & \textbf{4.91} $\pm$ \textbf{0.12} & $11.26 \pm 0.10$ & $184.2 \pm 3.5$ & $4.00 \pm 0.01$ \\
\midrule
TSFlow (S4)~\citep{kollovieh2025tsflow}    & $2.25 \pm 0.03$ (1) & $6.0 \pm 0.1$ (6) & $11.6 \pm 0.3$ (6) & $320 \pm 10$ (6) & $11.0 \pm 5.0$ (6) \\
Raw                                        & $2.31 \pm 0.02$ (6) & $5.13 \pm 0.09$ (1) & \textbf{10.0} $\pm$ \textbf{0.2} (1) & $181 \pm 3$ (5) & $4.04 \pm 0.08$ (3) \\
Fourier                                    & $2.27 \pm 0.03$ (3) & $5.1 \pm 0.1$ (1) & $10.1 \pm 0.2$ (1) & $176 \pm 2$ (2) & \textbf{3.95} $\pm$ \textbf{0.04} (1) \\
DCT                                        & $2.25 \pm 0.04$ (1) & $5.1 \pm 0.1$ (1) & \textbf{10.0} $\pm$ \textbf{0.1} (1) & \textbf{174} $\pm$ \textbf{3} (1) & $4.02 \pm 0.05$ (3) \\
PCA                                        & $2.26 \pm 0.02$ (1) & \textbf{5.0} $\pm$ \textbf{0.1} (1) & \textbf{10.0} $\pm$ \textbf{0.1} (1) & $178 \pm 2$ (4) & $4.00 \pm 0.06$ (3) \\
\textbf{MercerFlow (Ours)}                 & $2.24 \pm 0.03$ (1) & $5.1 \pm 0.1$ (1) & \underline{$10.1 \pm 0.2$} (1) & \textbf{174} $\pm$ \textbf{3} (1) & \underline{$3.97 \pm 0.05$} (1) \\
\bottomrule
\end{tabular}%
}
\end{small}
\end{table}

\subsection{Sensitivity to Extended Forecasting Horizons}
\label{app:extended-horizons}

To evaluate model resilience across extended horizons, we expand the forecast window from the default $L_{\mathrm{f}} = 192$ to $L_{\mathrm{f}} \in \{336, 720\}$ while keeping the context length fixed at $L_{\mathrm{p}} = 192$ (resulting in total sequence lengths of $L = 528$ and $L = 912$). Table~\ref{tab:horizon_sensitivity} provides the full test CRPS and MAE comparisons across all five benchmarks.

The results reveal distinct structural behaviors across datasets. On smooth series with long temporal correlation scales, such as \texttt{ETTh1} ($\ell = 336$), ambient observation-space modeling (\textsc{Raw}) remains effective. However, on complex, multi-frequency series such as \texttt{Weather} and \texttt{Electricity} at the longest horizon ($L_{\mathrm{f}} = 720$), predicting directly in observation space incurs severe error accumulation and variance inflation (e.g., \textsc{Raw} MAE on \texttt{Electricity} degrades to $206.9 \pm 15.6$, compared to $175.7 \pm 5.6$ for \textsc{DCT}). In contrast, orthogonal latent codecs constrain error compounding and maintain calibration even as the window length quadruples, demonstrating that compact MLP architectures operating in decorrelated coordinates scale reliably without requiring sequential recurrence.

\begin{table}[t]
\centering
\small
\setlength{\tabcolsep}{5pt}
\caption{Sensitivity to extended forecasting horizons ($L_{\mathrm{p}} = 192$). Bold: best; underline: second best.}
\label{tab:horizon_sensitivity}
\vskip 0.05in
\begin{tabular}{ccccccc}
\toprule
$L_{\mathrm{f}}$ & Method & ETTh1 & ETTh2 & Weather & Electricity & Traffic$^\dagger$ \\
\midrule
\multicolumn{7}{l}{\textbf{CRPS} ($\times 10^{-2}$, $\downarrow$)} \\
\midrule
\multirow{5}{*}{336} 
 & Raw        & \textbf{17.77} $\pm$ \textbf{0.23} & 13.07 $\pm$ 0.33                   & 3.98 $\pm$ 0.05                    & \textbf{4.43} $\pm$ \textbf{0.09} & \underline{10.36 $\pm$ 0.20} \\
 & Fourier    & 18.07 $\pm$ 0.49                   & 12.91 $\pm$ 0.19                   & \textbf{3.82} $\pm$ \textbf{0.02}  & 4.51 $\pm$ 0.10                   & 10.41 $\pm$ 0.15 \\
 & DCT        & \underline{17.81 $\pm$ 0.67}       & \underline{12.65 $\pm$ 0.31}       & \textbf{3.82} $\pm$ \textbf{0.02}  & \underline{4.49 $\pm$ 0.09}       & 10.45 $\pm$ 0.15 \\
 & PCA        & 17.93 $\pm$ 0.49                   & \textbf{12.51} $\pm$ \textbf{0.30} & 3.83 $\pm$ 0.03                    & \underline{4.49 $\pm$ 0.08}       & \textbf{10.29} $\pm$ \textbf{0.16} \\
 & MercerFlow & 17.82 $\pm$ 0.49                   & 12.66 $\pm$ 0.36                   & \textbf{3.82} $\pm$ \textbf{0.03}  & 4.52 $\pm$ 0.10                   & 10.41 $\pm$ 0.21 \\
\cmidrule(lr){1-7}
\multirow{5}{*}{720} 
 & Raw        & 18.61 $\pm$ 0.43                   & 11.28 $\pm$ 0.27                   & 5.13 $\pm$ 0.08                    & 4.98 $\pm$ 0.36                   & 12.06 $\pm$ 0.04 \\
 & Fourier    & \textbf{18.48} $\pm$ \textbf{0.58} & \textbf{10.35} $\pm$ \textbf{0.13} & \underline{4.78 $\pm$ 0.07}       & \underline{4.35 $\pm$ 0.12}       & 12.03 $\pm$ 0.07 \\
 & DCT        & \underline{18.52 $\pm$ 0.77}       & 11.52 $\pm$ 0.55                   & 4.79 $\pm$ 0.03                    & \textbf{4.33} $\pm$ \textbf{0.13} & \textbf{11.94} $\pm$ \textbf{0.07} \\
 & PCA        & 18.77 $\pm$ 0.87                   & \underline{10.99 $\pm$ 0.54}       & \textbf{4.76} $\pm$ \textbf{0.05}  & 4.43 $\pm$ 0.11                   & \underline{11.99 $\pm$ 0.13} \\
 & MercerFlow & 18.86 $\pm$ 0.76                   & 11.22 $\pm$ 0.54                   & 4.79 $\pm$ 0.07                    & 4.58 $\pm$ 0.21                   & 12.06 $\pm$ 0.10 \\
\midrule
\multicolumn{7}{l}{\textbf{MAE} ($\downarrow$)} \\
\midrule
\multirow{5}{*}{336} 
 & Raw        & \textbf{2.18} $\pm$ \textbf{0.03}  & 5.19 $\pm$ 0.12                   & 13.25 $\pm$ 0.29                   & \textbf{187.6} $\pm$ \textbf{3.9} & \underline{4.70 $\pm$ 0.10} \\
 & Fourier    & 2.20 $\pm$ 0.05                   & 5.12 $\pm$ 0.06                   & 13.11 $\pm$ 0.14                   & \underline{190.5 $\pm$ 4.0}       & \underline{4.70 $\pm$ 0.10} \\
 & DCT        & \textbf{2.18} $\pm$ \textbf{0.09}  & \underline{5.00 $\pm$ 0.12}       & \textbf{13.06} $\pm$ \textbf{0.18} & 190.6 $\pm$ 3.7                   & \underline{4.70 $\pm$ 0.10} \\
 & PCA        & 2.20 $\pm$ 0.06                   & \textbf{4.94} $\pm$ \textbf{0.13} & 13.08 $\pm$ 0.19                   & \underline{190.5 $\pm$ 3.7}       & \textbf{4.60} $\pm$ \textbf{0.10} \\
 & MercerFlow & \underline{2.19 $\pm$ 0.06}       & 5.01 $\pm$ 0.15                   & \underline{13.07 $\pm$ 0.15}       & 191.3 $\pm$ 4.2                   & \underline{4.70 $\pm$ 0.10} \\
\cmidrule(lr){1-7}
\multirow{5}{*}{720} 
 & Raw        & \textbf{2.38} $\pm$ \textbf{0.07}  & 4.50 $\pm$ 0.09                   & 15.37 $\pm$ 0.38                   & 206.9 $\pm$ 15.6                  & 5.30 $\pm$ 0.00 \\
 & Fourier    & \underline{2.40 $\pm$ 0.10}       & \textbf{4.20} $\pm$ \textbf{0.07} & \textbf{14.33} $\pm$ \textbf{0.21} & \underline{176.2 $\pm$ 6.6}       & 5.30 $\pm$ 0.00 \\
 & DCT        & 2.42 $\pm$ 0.13                   & 4.65 $\pm$ 0.23                   & 14.60 $\pm$ 0.25                   & \textbf{175.7} $\pm$ \textbf{5.6} & \textbf{5.20} $\pm$ \textbf{0.00} \\
 & PCA        & 2.46 $\pm$ 0.14                   & \underline{4.47 $\pm$ 0.21}       & \underline{14.51 $\pm$ 0.29}       & 182.0 $\pm$ 5.3                   & 5.30 $\pm$ 0.01 \\
 & MercerFlow & 2.48 $\pm$ 0.13                   & 4.54 $\pm$ 0.24                   & 14.63 $\pm$ 0.26                   & 189.1 $\pm$ 9.6                   & 5.30 $\pm$ 0.01 \\
\bottomrule
\multicolumn{7}{l}{\scriptsize $^\dagger$ Traffic MAE is reported in $\times 10^{-3}$.}
\end{tabular}
\end{table}       




\section{Optimizer Dynamics and Stability}
\label{app:high-lr}

Increasing the AdamW learning rate from the default $\eta = 10^{-3}$ to $\eta = 10^{-1}$ subjects the first-layer curvature to extreme stress under the shared OU prior.
Table~\ref{tab:joint_lr01} reports test CRPS at a matched mid-run evaluation checkpoint across all five benchmarks (seeds $42$--$44$). To ensure fair comparison, the shared evaluation checkpoint corresponds to the median of validation-best epochs among decorrelating codecs for each dataset (ETTh1 at epoch $40$, ETTh2 at $40$, Weather at $10$, Electricity at $20$, and Traffic at $30$).


\begin{table}[t]
\centering
\small
\caption{Test CRPS ($\times 10^{-2}$, $\downarrow$) at AdamW learning rate $\eta = 10^{-1}$ (default is $10^{-3}$), mean $\pm$ std over seeds $42$--$44$. Evaluated at matched mid-run EMA checkpoints $E^*$. Bold: within one std of the best.}
\label{tab:joint_lr01}
\vskip 0.05in
\setlength{\tabcolsep}{4pt}
\begin{tabular}{lccccc}
\toprule
Dataset & Mercer (Ours) & Fourier & DCT & PCA & Raw \\
\midrule
\texttt{ETTh1}       & $17.99 \pm 0.44$          & $18.17 \pm 0.55$          & $18.51 \pm 0.33$          & $\mathbf{17.81 \pm 0.00}$ & $122.42 \pm 80.10$ \\
\texttt{ETTh2}       & $\mathbf{13.05 \pm 0.56}$ & $13.27 \pm 0.06$          & $14.00 \pm 1.13$          & $\mathbf{12.92 \pm 0.20}$ & $15.18 \pm 0.69$ \\
\texttt{Weather}     & $\mathbf{1.70 \pm 0.01}$  & $\mathbf{1.71 \pm 0.01}$  & $\mathbf{1.71 \pm 0.01}$  & $\mathbf{1.70 \pm 0.01}$  & $\mathbf{1.71 \pm 0.06}$ \\
\texttt{Electricity} & $\mathbf{4.34 \pm 0.06}$  & $\mathbf{4.33 \pm 0.06}$  & $\mathbf{4.34 \pm 0.04}$  & $\mathbf{4.39 \pm 0.05}$  & $6.53 \pm 0.35$ \\
\texttt{Traffic}     & $\mathbf{10.65 \pm 0.36}$ & $\mathbf{10.80 \pm 0.86}$ & $11.16 \pm 0.26$          & $\mathbf{10.65 \pm 0.41}$ & $24.62 \pm 7.17$ \\
\bottomrule
\end{tabular}
\end{table}

Under this severe learning rate regime, the unrotated Raw baseline---whose first-layer condition number $\kappa(DAD)$ is two to four orders of magnitude larger than that of the orthonormal codecs---catastrophically collapses on the harder benchmarks (CRPS explodes to $1.22$ on ETTh1, $0.065$ on Electricity, and $0.246$ on Traffic).
In stark contrast, all decorrelating codecs (\textsc{MercerFlow}, Fourier, DCT, PCA) preserve stable descent trajectories, avoiding numerical instability and remaining close to their default-learning-rate performance.

\begin{figure}[t]
\centering
\includegraphics[width=0.85\linewidth]{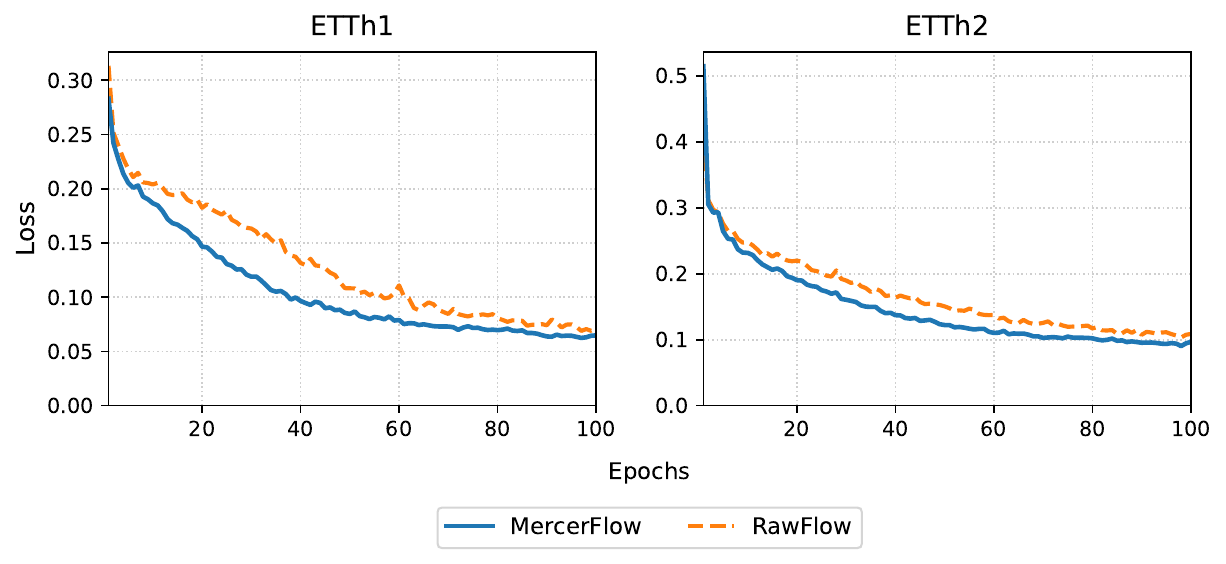}
\caption{Training loss trajectories on ETTh1 and ETTh2 at an elevated learning rate ($\eta = 10^{-2}$). \textsc{MercerFlow} (solid blue) converges faster and strictly monotonically. In contrast, the unrotated baseline (\textsc{RawFlow}, dashed orange) suffers from optimization friction and visible instability (e.g., the loss spike around epoch $60$ on ETTh1), demonstrating the empirical benefits of first-layer preconditioning.}
\label{fig:loss_curves_app}
\end{figure}

To visualize optimization stability, Figure~\ref{fig:loss_curves_app} plots the training loss trajectories at an elevated learning rate ($\eta = 10^{-2}$). \textsc{MercerFlow} converges rapidly and strictly monotonically. In contrast, the unrotated baseline (Raw, labelled \textsc{RawFlow} in the legend) exhibits pronounced optimisation friction and overt instability (e.g., a severe gradient-induced loss spike around epoch $60$ on ETTh1), visually confirming the critical role of first-layer preconditioning.

\section{Extended Ablations}
\label{app:extended-ablations}

\subsection{Synthetic Ornstein--Uhlenbeck Comparison}
\label{app:synthetic-ou}

Evaluating models on synthetic Ornstein--Uhlenbeck (OU) sample paths provides a strictly controlled testbed with a known data-generating kernel. Under this setting, matched and mismatched linear codecs are analytically well-defined, allowing us to evaluate the impact of first-layer conditioning in isolation from real-world confounding factors.

Synthetic trajectories are generated with context length $L_p = 32$, forecast horizon $L_f = 32$ (total window $L = 64$), and stationary length scale $\ell = 32$ (corresponding to one $e$-fold decay across the boundary between context and horizon). The observation-space GP prior and the Mercer basis share this identical ground-truth kernel. Fourier, DCT, and the unrotated Raw baseline employ the exact same AdaLN MLP backbone (width $64$, $2$ hidden layers) and the same conditional GP prior; only the linear coordinate transformation $\Phi \in \mathbb{R}^{L \times L}$ differs. PCA is omitted in this setting as the ground-truth kernel is known analytically. All models are trained with AdamW at $\eta = 10^{-3}$ for $2{,}000$ steps (batch size $64$, random seeds $0$--$2$).

Table~\ref{tab:synthetic-ou} reports the Jacobi condition number $\kappa(DAD)$ evaluated on the centred prior population ($\Phi^\top K \Phi$) and on empirical conditional CFM interpolants $c_t$, along with test CRPS ($20$ Euler integration steps, $32$ trajectories, $256$ evaluation windows).

\paragraph{Matched OU Prior.}
On the centred prior covariance, \textsc{MercerFlow} analytically achieves optimal conditioning ($\kappa(DAD) = 1.00$). DCT closely follows ($\kappa(DAD) = 1.82$), Fourier sits in the tens ($24.9$), and the unrotated Raw baseline suffers from severe ill-conditioning ($\kappa(DAD) = 1{,}501.0$).
On empirical conditional interpolants, conditioning degrades across all codecs due to non-zero context conditioning, but preserves the identical relative ordering: Mercer ($98.5$) and DCT ($105.0$) remain around $10^2$, Fourier sits at $654.0$, and Raw reaches $3{,}515.0$.
Under AdamW, all four codecs tie on test CRPS within $0.2\%$ (DCT $0.6062 \pm 0.0024$, Mercer $0.6063 \pm 0.0025$, Fourier $0.6064 \pm 0.0023$, Raw $0.6069 \pm 0.0030$). This flat scoreboard reflects the synthetic ``easy'' regime: because the prior matches the data-generating process exactly, the prior conditional mean already explains almost all predictive variance, leaving minimal residual dynamics for the flow to resolve. Nevertheless, the geometric separation remains stark in the condition number columns.

\paragraph{Mismatched Prior Length Scale (Wrong-$\ell$ Stress Test).}
To examine sensitivity to kernel misspecification, we train a mismatched \textsc{MercerFlow} arm where the GP model and Mercer basis assume $\ell = 320$ ($10\times$ larger than the ground-truth data length scale $\ell = 32$). This mismatch increases the prior $\kappa(DAD)$ from $1.00$ to $1.71$ and the interpolant $\kappa(DAD)$ from $98.5$ to $192.0$, resulting in a mild CRPS degradation ($0.6063 \to 0.6125 \pm 0.0057$). Crucially, this performance gap ($0.0062$) remains within seed standard deviation, confirming that \textsc{MercerFlow} maintains stable optimization even under moderately misspecified prior parameters.

\paragraph{SGD Equivariance Control.}
Training with Nesterov SGD (momentum $0.9$, learning rate $10^{-3}$, matched step budget) completely eliminates all performance discrepancies: Mercer, Fourier, and Raw achieve identical test CRPS ($0.6033 \pm 0.0034$), with a relative gap from Raw of exactly $0.00\%$. This empirical result provides definitive proof of Proposition~1, demonstrating that coordinate representations have zero effect under rotation-equivariant optimization.


\begin{table}[hbt!]
\centering
\caption{Synthetic OU comparison (context/horizon $32/32$, data length scale $\ell=32$).
Matched codecs share the OU prior at $\ell=32$; the wrong-$\ell$ arm uses a Mercer/GP model at $\ell=320$.
AdaLN MLP ($64$ hidden, $2$ layers), AdamW $\eta = 10^{-3}$, $2{,}000$ steps, seeds~$0$--$2$ (mean $\pm$ std).
$\kappa(DAD)$ prior is computed on the population covariance $\Phi^\top K\Phi$; interpolant $\kappa(DAD)$ is empirical on conditional CFM states $c_t$.
SGD control: Nesterov momentum $0.9$, matched compute budget; gaps are $100 \times (\mathrm{CRPS}/\mathrm{CRPS}_{\mathrm{Raw}} - 1)$ (mean over seeds).
Bold indicates values within one standard deviation of the best.}
\label{tab:synthetic-ou}
\vskip 0.05in
\begin{small}
\begin{tabular}{lcccc}
\toprule
Codec / arm & $\kappa(DAD)$ prior & $\kappa(DAD)$ interp. & Test CRPS ($\downarrow$) & SGD gap (\%) \\
\midrule
Mercer ($\ell=32$)  & $1.00$    & $98.5$    & $\mathbf{0.6063 \pm 0.0025}$ & $0.00$ \\
Fourier ($\ell=32$) & $24.90$   & $654.0$   & $\mathbf{0.6064 \pm 0.0023}$ & $0.00$ \\
DCT ($\ell=32$)     & $1.82$    & $105.0$   & $\mathbf{0.6062 \pm 0.0024}$ & $0.00$ \\
Raw ($\ell=32$)     & $1{,}501.0$ & $3{,}515.0$ & $\mathbf{0.6069 \pm 0.0030}$ & $0.00$ \\
\midrule
Mercer ($\ell=320$) & $1.71$    & $192.0$   & $0.6125 \pm 0.0057$ & --- \\
\bottomrule
\end{tabular}
\end{small}
\end{table}

\subsection{Sampling Efficiency: Flow Matching vs. Diffusion}
\label{app:diffusion-comparison}

Conditional flow matching generates straight probability paths by pairing source prior draws directly with data targets, avoiding the stochastic Brownian simulation inherent to diffusion.
Table~\ref{tab:diffusion_nfe} benchmarks \textsc{MercerFlow} against an equivalent score-based diffusion model formulated in the identical Mercer latent space (\textsc{MercerDiff}) across varying Numbers of Function Evaluations (NFE) at forecast horizon $L_f = 192$.

While \textsc{MercerDiff} exhibits severe discretization errors at low step budgets and remains suboptimal even at $\mathrm{NFE} = 128$ ($0.47$ on ETTh1), \textsc{MercerFlow} fully saturates its predictive distribution within merely $4$--$10$ Euler integration steps ($0.199$), providing a massive practical acceleration during inference.

\begin{table}[hbt!]
\centering
\caption{Inference sampling efficiency: Test CRPS ($\downarrow$, mean $\pm$ std) across varying Numbers of Function Evaluations (NFE) at horizon $L_{\mathrm{f}} = 192$. Due to straight probability paths, \textsc{MercerFlow} saturates within $4$--$10$ steps, whereas diffusion (\textsc{MercerDiff}) suffers from severe discretization error. Bold: best per column.}
\label{tab:diffusion_nfe}
\vskip 0.05in
\small
\setlength{\tabcolsep}{0pt}
\begin{tabular*}{\textwidth}{@{\extracolsep{\fill}} llccccc}
\toprule
& & \multicolumn{5}{c}{\textbf{NFE (Euler Steps)}} \\
\cmidrule(lr){3-7}
Dataset & Method & 4 & 10 & 20 & 50 & 128 \\
\midrule
\multirow{2}{*}{\texttt{ETTh1}} 
 & MercerDiff & $19.03 \pm 0.07$           & $1.04 \pm 0.06$            & $0.56 \pm 0.07$            & $0.51 \pm 0.06$   & $0.47 \pm 0.03$ \\
 & MercerFlow & $\mathbf{0.200 \pm 0.017}$ & $\mathbf{0.199 \pm 0.017}$ & $\mathbf{0.199 \pm 0.017}$ & \textemdash       & \textemdash \\
\midrule
\multirow{2}{*}{\texttt{ETTh2}} 
 & MercerDiff & $7.78 \pm 0.02$            & $0.47 \pm 0.01$            & $0.211 \pm 0.005$          & $0.176 \pm 0.005$ & $0.177 \pm 0.003$ \\
 & MercerFlow & $\mathbf{0.137 \pm 0.005}$ & $\mathbf{0.136 \pm 0.006}$ & $\mathbf{0.136 \pm 0.007}$ & \textemdash       & \textemdash \\
\bottomrule
\end{tabular*}
\end{table}      
\subsection{Sensitivity to Prior Diagonal Jitter}
\label{app:jitter-ablation}

The regularising diagonal jitter $\varepsilon I_L$ added to the Gaussian process covariance matrix $K$ serves two distinct purposes: it ensures numerical stability during matrix inversion in the conditional prior (Equation~\ref{eq:gp-prior}) and sets the diagonal regularisation floor for the Mercer eigenspace.
A minimal jitter level ($\varepsilon \le 10^{-3}$) preserves exact alignment with the continuous analytical kernel, whereas excessive jitter forces the prior covariance toward isotropic white noise ($K \approx I_L$).
Table~\ref{tab:jitter_ablation} systematically evaluates the sensitivity of first-layer conditioning $\kappa(DAD)$ (measured on conditional interpolants $\mathbf{c}_t$) and test CRPS across jitter values $\varepsilon \in [10^{-4}, 1.0]$ on \texttt{ETTh1} and \texttt{ETTh2}.

\paragraph{Stability Under Standard Regularisation ($\varepsilon \le 10^{-2}$).}
For standard jitter scales ($\varepsilon \in [10^{-4}, 10^{-2}]$), \textsc{MercerFlow} maintains tightly bounded, well-conditioned first-layer geometry ($\kappa(DAD) \le 30.6$ on \texttt{ETTh1} and $\le 95.3$ on \texttt{ETTh2}), yielding stable predictive performance (CRPS $0.186$--$0.188$ on \texttt{ETTh1}).
Conversely, diagonal jitter fails to resolve the conditioning pathology of the unrotated \textsc{Raw} baseline, whose condition number remains several orders of magnitude higher ($\kappa(DAD) \sim 10^5$--$10^6$) across all evaluated jitter scales.
On the simple \texttt{ETTh2} benchmark, where the GP prior mean accounts for the dominant dynamics, CRPS remains virtually invariant to the choice of codec ($0.131$--$0.135$).

\paragraph{Breakdown Under Extreme Jitter ($\varepsilon \ge 0.1$).}
At extreme jitter levels ($\varepsilon \ge 0.1$), the kernel's continuous temporal correlation is drowned out by artificial diagonal dampening ($K \to I_L$).
This distorts the kernel's eigenspace, driving Mercer's condition number up to $3.64 \times 10^3$ on \texttt{ETTh2}.
More critically, because the prior degenerates toward uninformative white noise, the flow loses its data-matched starting state, causing pronounced CRPS degradation on the harder \texttt{ETTh1} benchmark ($0.188 \to 0.255$ for Mercer and $0.192 \to 0.256$ for \textsc{Raw}).
These results show that \textsc{MercerFlow}'s effectiveness stems from the kernel's authentic analytical structure rather than artificial diagonal conditioning.

\begin{table}[t]
\centering
\caption{Sensitivity to GP prior diagonal jitter $\varepsilon$ on \texttt{ETTh1} and \texttt{ETTh2} ($L = 384$). $\kappa(DAD)$ is evaluated on conditional interpolants $\mathbf{c}_t$. Standard regularisation ($\varepsilon \le 10^{-2}$) preserves well-conditioned representations; extreme jitter ($\varepsilon \ge 0.1$) collapses the prior toward white noise ($K \approx I_L$), degrading predictive accuracy. Bold: best per column within each dataset block.}
\label{tab:jitter_ablation}
\vskip 0.05in
\small
\resizebox{\linewidth}{!}{%
\setlength{\tabcolsep}{6pt}
\begin{tabular}{lllccccc}
\toprule
& & & \multicolumn{5}{c}{\textbf{Diagonal Jitter} $\varepsilon$} \\
\cmidrule(lr){4-8}
Metric & Dataset & Codec & $10^{-4}$ & $10^{-3}$ & $10^{-2}$ & $10^{-1}$ & $1.0$ \\
\midrule
\multirow{2}{*}{\shortstack[l]{$\kappa(DAD)$\\($\mathbf{c}_t$, $\downarrow$)}} 
 & \multirow{2}{*}{\texttt{ETTh1}} 
   & Mercer & \textbf{23.55} & \textbf{30.64} & \textbf{29.37} & \textbf{47.03} & \textbf{334.67} \\
 & & Raw    & $2.01 \times 10^5$ & $3.02 \times 10^5$ & $1.85 \times 10^5$ & $1.66 \times 10^5$ & $1.44 \times 10^5$ \\
\cmidrule(lr){2-8}
 & \multirow{2}{*}{\texttt{ETTh2}} 
   & Mercer & \textbf{62.14} & \textbf{74.75} & \textbf{95.27} & \textbf{524.26} & $\mathbf{3.64 \times 10^3}$ \\
 & & Raw    & $4.07 \times 10^5$ & $1.22 \times 10^6$ & $1.20 \times 10^6$ & $1.17 \times 10^6$ & $1.01 \times 10^6$ \\
\midrule
\multirow{2}{*}{\shortstack[l]{CRPS\\($\downarrow$)}} 
 & \multirow{2}{*}{\texttt{ETTh1}} 
   & Mercer & \textbf{0.188} & \textbf{0.186} & \textbf{0.187} & \textbf{0.195} & \textbf{0.255} \\
 & & Raw    & 0.192 & 0.187 & \textbf{0.187} & 0.196 & 0.256 \\
\cmidrule(lr){2-8}
 & \multirow{2}{*}{\texttt{ETTh2}} 
   & Mercer & 0.133 & \textbf{0.132} & 0.135 & 0.132 & \textbf{0.136} \\
 & & Raw    & \textbf{0.131} & 0.135 & \textbf{0.134} & \textbf{0.131} & \textbf{0.136} \\
\bottomrule
\end{tabular}%
}
\end{table}           

\end{document}